\documentclass{article}

\PassOptionsToPackage{bookmarks=true}{hyperref}
\usepackage[preprint,nonatbib]{corl_2026} %

\usepackage{amsmath,amsfonts,bm}

\def\eqref#1{equation~\ref{#1}}

\def\1{\bm{1}}

\DeclareMathAlphabet{\mathsfit}{\encodingdefault}{\sfdefault}{m}{sl}
\SetMathAlphabet{\mathsfit}{bold}{\encodingdefault}{\sfdefault}{bx}{n}

\DeclareMathOperator{\Tr}{Tr}

\usepackage[ruled,vlined,linesnumbered,noend]{algorithm2e}
\usepackage{xcolor}
\usepackage{xcolor}
\usepackage[most]{tcolorbox}
\usepackage{enumitem}
\usepackage{multicol}
\usepackage{wrapfig}
\usepackage{times}
\usepackage{amsmath}
\usepackage{amssymb}
\usepackage{siunitx}
\usepackage{gensymb}
\usepackage{booktabs}
\usepackage{multirow}
\usepackage{authblk}
\usepackage{pifont}
\usepackage{soul}
\usepackage{array}
\usepackage{adjustbox}
\usepackage{subcaption}
\usepackage{amsmath}
\usepackage{threeparttable}
\usepackage{pifont} %

\usepackage[numbers,sort&compress]{natbib}
\usepackage{multicol}
\usepackage[bookmarks=true]{hyperref}
\usepackage{float}

\tcbuselibrary{skins} \usepackage{enumitem}

\SetKwInput{KwInput}{Inputs}
\SetKwInput{KwOutput}{Outputs}
\SetKwInput{KwParams}{Hyperparameters}

\SetCommentSty{mycommfont}

\SetAlgoInsideSkip{smallskip}
\SetAlgoNlRelativeSize{-1}
\SetAlCapNameSty{textsc}
\SetAlCapSty{}
\DontPrintSemicolon
\RestyleAlgo{ruled}

\definecolor{my_orange}{HTML}{FF8A00}
\definecolor{my_blue}{HTML}{4363D8}
\definecolor{my_green}{HTML}{3CB44B}

\newif\ifanonymous
\anonymoustrue %
\newcommand{\projectsite}{\ifanonymous transformer-transformer-anonymous.github.io\else transformer-transformer.github.io\fi}
\newcommand{\projectlink}{\href{https://\projectsite}{\projectsite}}
\newcommand{\paperauthors}{\ifanonymous Anonymous Author(s)\else Huy Ha$^{1,2}$, Karen Liu$^{1}$, Shuran Song$^{1,2}$ \\ $^1$Stanford University \quad $^2$Columbia University\fi}
\newif\ifsqueeze
\squeezetrue %
\newcommand{\vsqueeze}[1]{\ifsqueeze\vspace{#1}\fi}
\newenvironment{maybewrap}[2]
  {\ifsqueeze\wrapfigure{#1}{#2}\else\figure[t]\centering\minipage{#2}\fi}
  {\ifsqueeze\endwrapfigure\else\endminipage\endfigure\fi}
\usepackage{needspace}
\newcommand{\wrapclearance}[1]{\ifsqueeze\else\needspace{#1\baselineskip}\fi}

\anonymousfalse %
\squeezefalse %
\ifsqueeze\else\fi

\ifanonymous\else
\hypersetup{
   pdfauthor={Huy Ha, Karen Liu, Shuran Song},
   pdftitle={Transformer Transformer: A Unified Model for Motion-Conditioned Robot Co-design},
   pdfsubject={Robotics},
}
\fi

\title{
   Transformer Transformer:
   A Unified Model for Motion-Conditioned Robot Co-design
}

\author{\paperauthors}

\begin{document}
\maketitle
\vsqueeze{-8mm}
\begin{figure}[h]
   \centering
   \includegraphics[width=\textwidth]{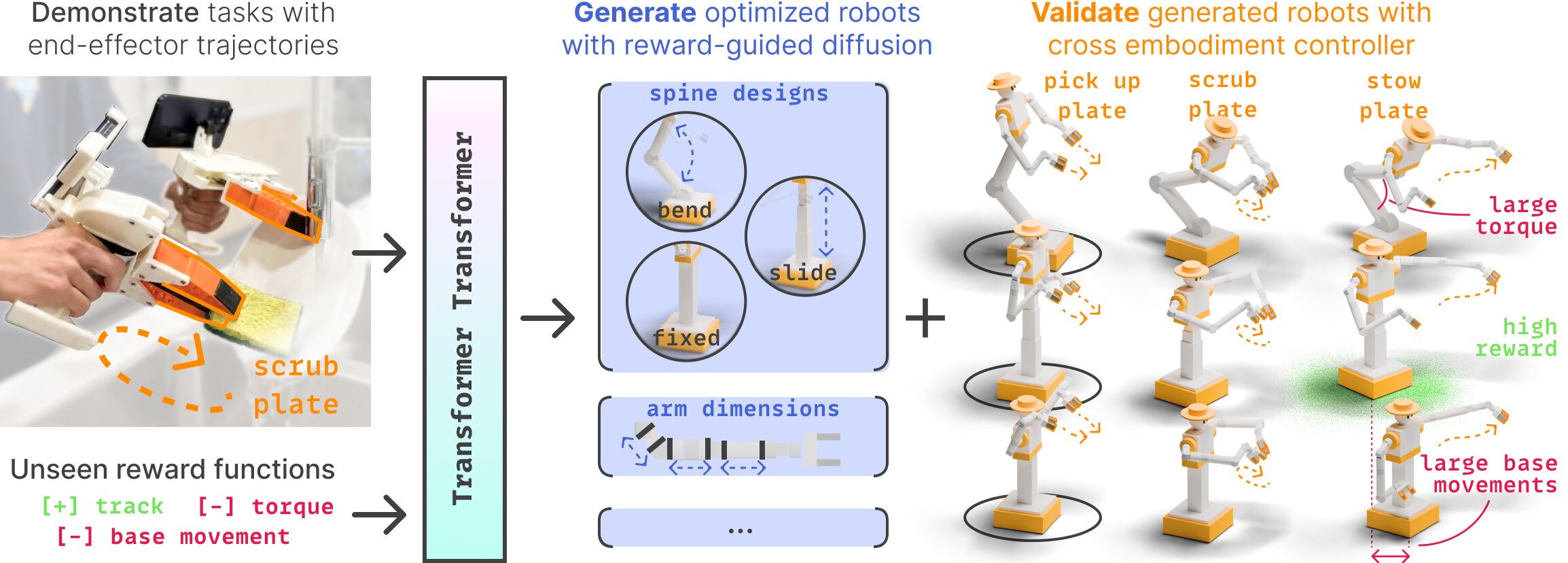}
   \caption{\textbf{
         Demonstrate, Generate, Validate.
      } From target end-effector motions and \emph{unseen} reward functions, Transformer Transformer \textcolor{my_blue}{\textbf{generates optimized robot}} designs that maximize rewards, then \textcolor{my_orange}{\textbf{validates generated robots}} by controlling them to track the given motions.
      Trained over robot embodiment, state, and action tokens, it unifies embodiment optimization and cross-embodiment control into one model, providing a one-stop shop for robot co-design from human demonstrations.
   }
   \vsqueeze{-2mm}
   \label{fig:teaser}
\end{figure}

\begin{abstract}
   An often overlooked factor of robot manipulation performance is the embodiment of the robot itself.
   Motivated by this problem, we study motion-conditioned robot co-design, where the goal is to generate complete robot designs that track target end-effector trajectories (from human demonstrations) while optimizing user-defined rewards. We introduce Transformer Transformer, a diffusion transformer trained on RoboTokens, a unified tokenization of robot embodiments, states, and actions.
   The same architecture can be used across embodiment spaces (e.g., wheeled bimanual, quadrupeds, humanoids) and use cases (embodiment generation, cross embodiment controller).
   Rather than overfitting to one reward function, Transformer Transformer is a dynamics model, whose reward-agnostic state and action predictions can be converted into reward-specific value predictions.
   These value predictions are used to steer embodiment diffusion towards high value robot designs, through a procedure we call Dynamics Self-Guidance.
   Experiments across multiple design spaces show zero-shot optimization of unseen rewards and trajectories, improving performance and runtime over the evolutionary baseline.
   Finally, we fabricated an optimized ALOHA design, which reduced tracking error by over $70\%$ compared to the original design.
   Check out \projectlink{} for summary/result videos.
\end{abstract}


\section{Introduction}

An often overlooked factor of manipulation performance is the embodiment of the robot itself.
What motions robots can effectively perform depends on its embodiment.
Motivated by this, we ask:
\begin{quote}
    \centering
    \vsqueeze{-2mm}
    \textit{
        What is the best robot embodiment
        for a given manipulation task?}
    \vsqueeze{-2mm}
\end{quote}
We formalize manipulation tasks as end-effector trajectories, a widely adopted task representation for manipulation policy learning~\cite{chi2024universal,liu2024maniwav,lin2024data,wu2024fast,seo2024legato,liu2025vitamin,xu2025dexumi,tao2025dexwild,zhu2025touch,lee2025manipforce,rayyan2025mv,xu2025exumi,gupta2025umionair}.
While these end-effector trajectories are embodiment-agnostic, their execution is not.
Instead, it critically depends on the tracking performance dictated by the robot embodiment.
As a result, policies trained on such data transfer unequally to different robots~\cite{gupta2025umionair}.
Seen this way, imperfect transfer is not only a limitation—it is also an opportunity.

In this work, we study \emph{motion-conditioned robot co-design}. %
Given (i) a set of target end-effector trajectories and (ii) a set of user-defined reward functions, our goal is to automatically generate \emph{complete} robot embodiments that achieve high rewards (Fig.~\ref{fig:teaser}).
Here, a complete robot embodiment means reasoning over large, heterogeneous design spaces spanning kinematics, geometry, and dynamics parameters.
Further, the user-specified rewards can be functions of both the robot's embodiment (e.g., size, mass) and dynamic behavior (e.g., joint velocity, tracking precision).

To tackle this problem, we propose \textbf{Transformer Transformer}%
\footnote{The first ``Transformer'' refers to shape-shifting robots; the second refers to the self-attention architecture.},
a DiT~\cite{peebles2023scalable} designed to be a one-stop shop for motion-conditioned robot co-design, enabled by the following contributions:
\begin{itemize}[leftmargin=3mm,itemsep=0pt]
    \vsqueeze{-2mm}
    \item \textbf{Unified Robot Representation:}
          We propose RoboTokens, a tokenization scheme that can represent any articulated robot's embodiment, states, and actions.
          Designed to flexibly store complete robot data in a consistent format, RoboTokens provides the foundational representational capability that enables a single model to span diverse robot embodiment spaces.
    \item \textbf{Unified Architecture:}
          Trained with different masked modeling schemes over RoboTokens, our model acts simultaneously as the generator, critic, and controller for co-design problems.
          By consolidating all co-design modules into a single neural network, our optimization pipeline is simple and readily GPU-parallelized.
    \item \textbf{Unified Dynamics Training Objective:}
          Instead of specializing on a reward-specific signal, our network is trained to model the general dynamics of diverse robot embodiments.
          To zero-shot optimize for user-specified reward functions at inference time, we convert our model's \emph{reward-agnostic} dynamics prediction into a \emph{reward-specific} score prediction.
          This predicted score is used to guide the model's embodiment diffusion process to produce high-value embodiments.
          \vsqueeze{-2mm}
\end{itemize}
Across three robot design spaces (i.e., fixed-base, quadruped, and bimanual mobile manipulator), we demonstrate that Transformer Transformer can zero-shot optimize user-specified end-effector trajectories and reward functions at inference time.
Conditioned on the robot embodiment, the same model can also be used as a cross-embodiment whole-body controller, directly usable for design validation.
Finally, we validate our model's predictions by fabricating an optimized ALOHA design on bimanual flinging for cloth unfolding, reducing tracking error by $73\%$ and maximum joint velocity by $30\%$.
Check out the \projectlink{} for more results.

\section{Method}
\label{sec:method}

\begin{wrapfigure}{r}{0.5\textwidth}
    \vsqueeze{-12mm}
    \includegraphics[width=\linewidth]{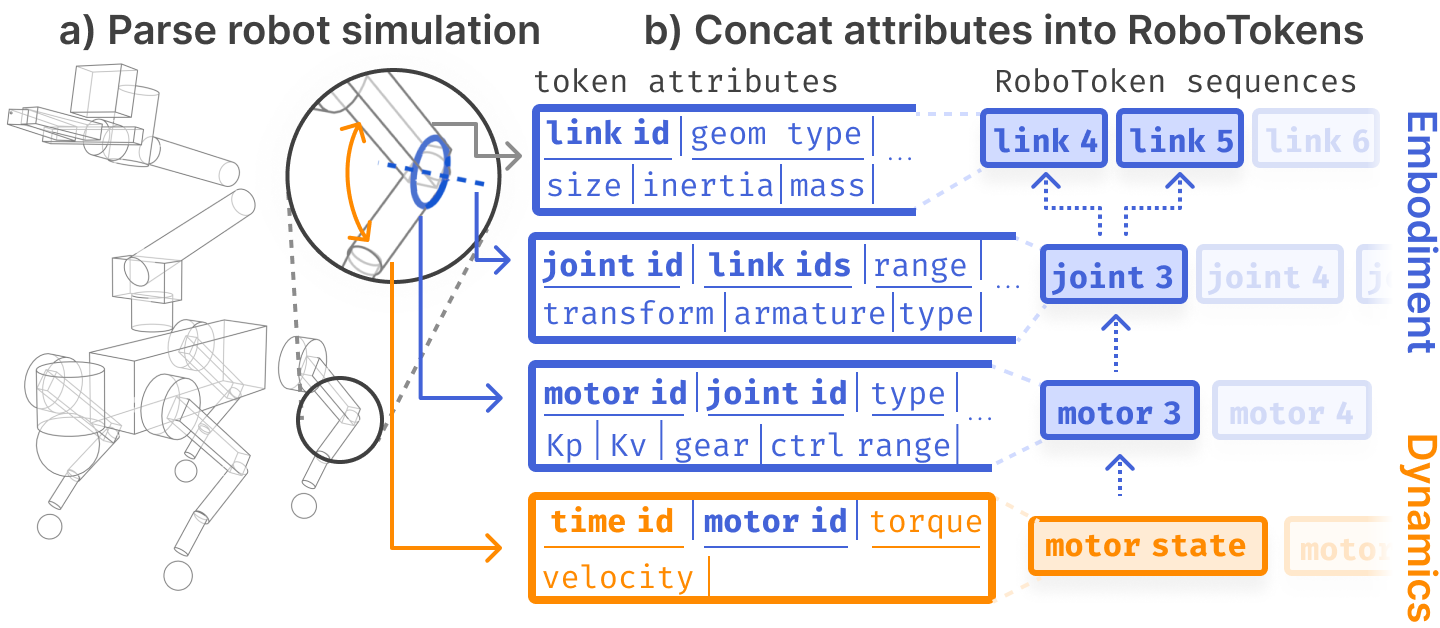}
    \vspace{-6mm}%
    \caption{
        \small
        \textbf{A Unified Robot Representation}.
        RoboToken is designed to be a complete, learning-ready token representation of rigid articulated robots.
        Our tokenizer can convert any robot description into a sequence of RoboTokens, consisting of time-invariant \textcolor{my_blue}{\textbf{embodiment}} tokens
        and time-varying \textcolor{my_orange}{\textbf{dynamics}} tokens.
    }
    \vspace{-3mm}%
    \label{fig:tokenization}
\end{wrapfigure}

\textbf{Overview:}
The overarching goal of Transformer Transformer is to model the joint distribution of robots' embodiment and dynamics (i.e., states and actions).
To enable this goal, we first describe RoboToken (\S~\ref{sec:robotoken}), a tokenization scheme that can represent the embodiments and dynamics from diverse robots.
Next, we will describe how to train a diffusion transformer over these tokens (\S~\ref{sec:arch:training}), along with its applications in motion-to-robot generation and cross-embodiment control policies.
To enable more efficient sampling, we describe dynamics self-guidance, a method for guiding the diffusion process to optimize unseen reward functions without training any additional network (\S~\ref{sec:selfguidance}).

\noindent{\textbf{Scope:}}
The RoboToken representation was designed to represent rigid articulated robots (same scope as MJCF) optimized for robot learning. It is currently limited to primitive-based geometry representation. Further, we focus on end-effector trajectories from UMI~\cite{chi2024universal} demonstrations as the manipulation task representation, which is widely adopted~\cite{chi2024universal,liu2024maniwav,lin2024data,wu2024fast,seo2024legato,liu2025vitamin,xu2025dexumi,tao2025dexwild,zhu2025touch,lee2025manipforce,rayyan2025mv,xu2025exumi}.
Finally, we aim to design a framework that can generalize to reward functions over the robot's embodiment, states, and controls.
Rewards involving other properties (e.g., structural strength, appearance) are out of scope.

\subsection{RoboToken: A Unifying Robot Representation}
\label{sec:robotoken}
\vsqueeze{-3mm}

\textbf{Complete.}
RoboToken includes all attributes of a robot's embodiment (time-invariant) and dynamics (time-varying).
The tokenizer converts any robot description into a set of five embodiment token types: links, fixed joints, sliding/rotating joints, ball joints, and motors (Fig.~\ref{fig:tokenization}, only three embodiment types shown for illustration purposes).
In addition, the tokenizer also converts each episode of states and actions into a set of four state token types (all but fixed joints), along with action tokens.
For instance, each link token contains the physical parameters of the link (primitive type, primitive size, inertia), while link state tokens include the link's pose at a given timestep.
Within each token type, all data attributes are concatenated into a learning-ready, continuous-valued vector.

\begin{wrapfigure}{r}{0.45\textwidth}
    \vsqueeze{-3mm}
    \centering
    \includegraphics[width=\linewidth]{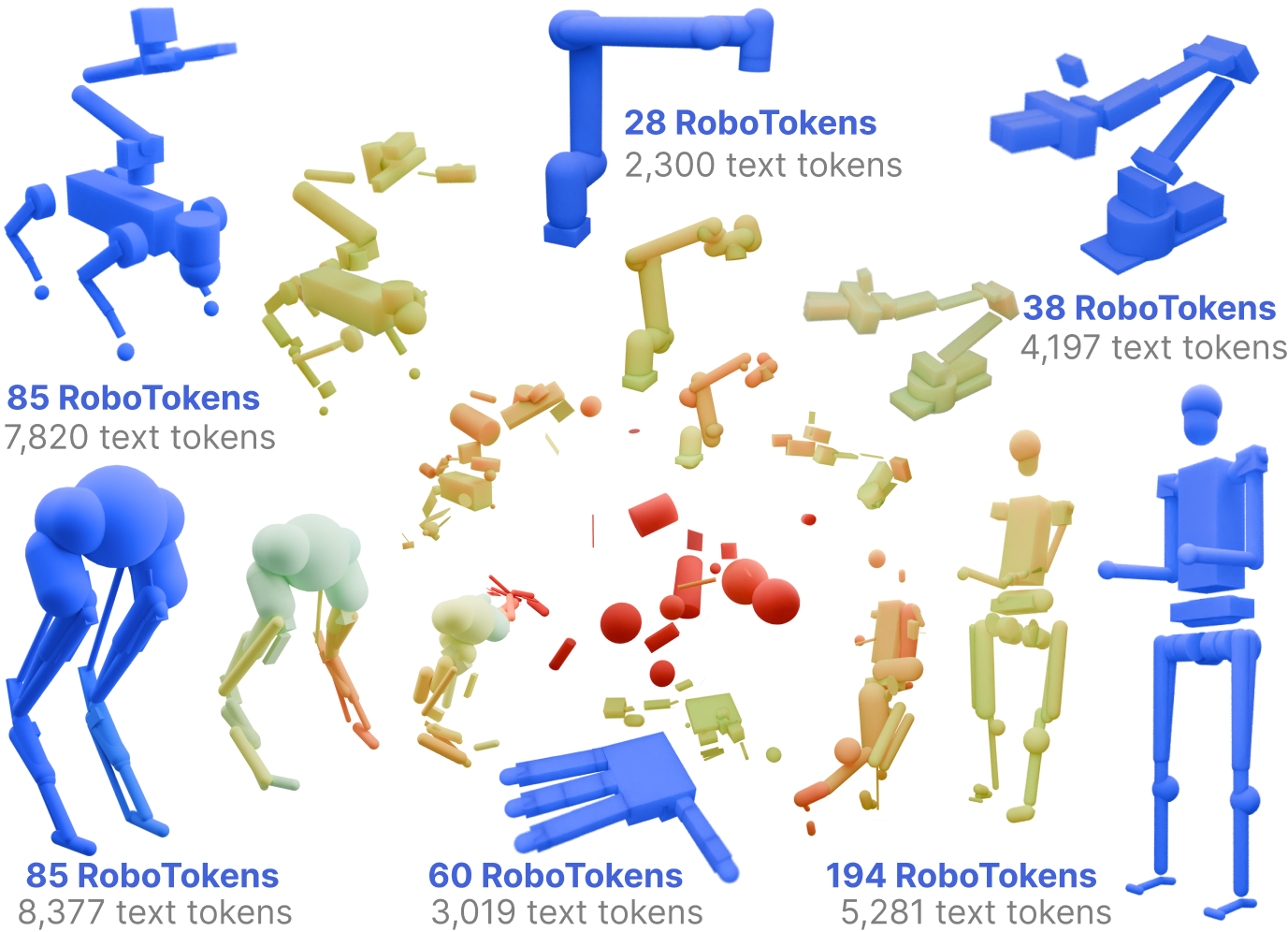}
    \vsqueeze{-6mm}
    \caption{
        \small
        \textbf{RoboToken Diffusion}.
        Transformer Transformer diffuses \textcolor{red}{noise} into \textcolor{my_blue}{RoboTokens (token count in blue)}, 27-110$\times$ more compact than \textcolor{gray}{MJCF text tokens (token count in grey)}.
    }
    \label{fig:robotokens}
    \vsqueeze{-7mm}
\end{wrapfigure}

\textbf{Flexible.}
To encompass diverse robot kinematics, RoboToken supports a variable number of embodiment tokens representing arbitrary connectivity.
Here, joints point to the two links they connect via two link IDs, while motors point to the joint they control with joint IDs (Fig.~\ref{fig:tokenization}, dotted arrows).
This supports passive joints when no motors point to them (e.g., Cassie's leg mechanism, Fig.~\ref{fig:robotokens}).
To point to each other, tokens include each other's IDs in their data attributes.
To support heterogeneous state and action spaces, we represent each link/joint/motor state at each timestep as its own token, pointing to the corresponding timestep and embodiment token.
These IDs are used later in training to learn positional embeddings for each ID type (Fig.~\ref{fig:architecture}).

\textbf{Consistent.}
Robot description formats like MJCF do not have strict conventions regarding transform handling, opting for ease of human editing instead of data representation consistency.
As such, directly learning over these text representations would force networks to learn redundant spatial offsets that are equivalent, introducing extra variance without extra information.
The RoboToken tokenizer enforces consistent transform conventions for representing geometry and dynamics information, automatically splitting inertias, and collapsing transforms in the preprocessing stage.

\textbf{Extensible.}
Our implementation of RoboToken makes it frictionless to support extra tokenization/detokenization logic for different robot tasks, embodiment, and dynamics data.
To illustrate this, we adapted RoboToken for the trajectory tracking task by adding target end-effector pose tokens to the RoboToken format (Fig.~\ref{fig:architecture}, ``Target Pose'' tokens).
These conditioning tokens aren't noised during training, and point to the end-effector ID that should track it and the timestep ID.

\textbf{Optimizability.}
While text-based formats like MJCF can be used for generation, they are difficult to \emph{optimize} over.
LLMs are autoregressive, categorical models that lack \emph{global controllability} (i.e., later tokens can't affect earlier ones) and \emph{differentiability} (due to sampling and detokenization).
In contrast, diffusing continuous-valued RoboToken has both these properties.

\vsqueeze{-3mm}
\subsection{Transformer Transformer: A Unifying Architecture}
\label{sec:arch:training}
\vsqueeze{-3mm}

\begin{figure*}[t]
    \centering
    \includegraphics[width=\linewidth]{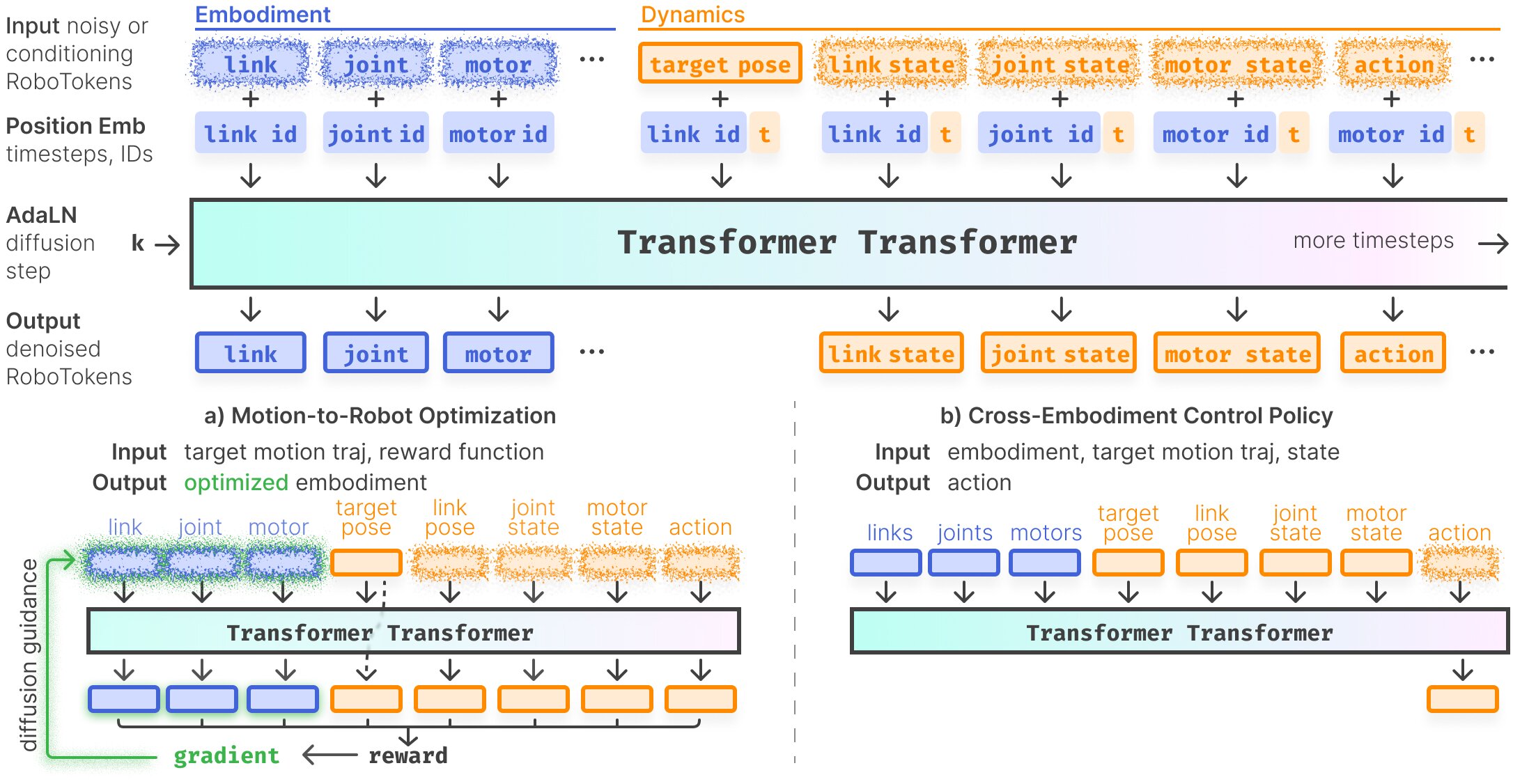}
    \caption{
        \small
        \textbf{Unified Architecture for Robot Co-Design}.
        Transformer Transformer is a diffusion transformer, trained to jointly diffuse \textcolor{my_blue}{\textbf{embodiment}} and \textcolor{my_orange}{\textbf{dynamics}} RoboTokens, where different conditioning in the same model enables
        (a) motion-to-robot optimization and (b) cross-embodiment control.
        At inference time, the model uses unseen rewards to turn its \textcolor{my_blue}{\textbf{embodiment}} and \textcolor{my_orange}{\textbf{dynamics}} predictions into a \textbf{reward} prediction, and incorporates its \textcolor{my_green}{\textbf{gradients}} back into the diffusion process.
    }
    \vsqueeze{-7mm}
    \label{fig:architecture}
\end{figure*}

To train a model over RoboTokens, we need an architecture that can model \textit{variable} sequence lengths of \textit{high-precision continuous-valued} data.
To achieve this, we use a diffusion transformer (DiT)-based architecture~\cite{peebles2023scalable} (Fig.~\ref{fig:architecture}).
Trained with different masked modeling schemes, our model can both generate robot designs and control them for motion-conditioned co-design, as follows:

\textbf{Motion-to-Robot Optimization.}
Given target motions and reward functions, the model must generate high-value robot designs.
Evaluating these reward functions often requires both information about the embodiment (e.g., mass penalty) and dynamics (e.g., torque penalty) while tracking the target motion (e.g., tracking reward).
Thus, conditioned on the target motion trajectory, our model \emph{simultaneously} diffuses embodiment and dynamics tokens, which gives the model everything it needs to assess newly generated designs.
Beyond speed, this non-autoregressive formulation also avoids the error accumulation inherent to autoregressive dynamics models, yielding more temporally coherent long-horizon predictions for design evaluation.
When trained on data from diverse robots, the model must learn long-range dependencies between robot design, states, and controls over extended time horizons.
Since the model also outputs action tokens, this optimization procedure also considers control.
We construct an embodiment generator biased toward high-value candidates by running $n$ parallel diffusion processes, using the specified reward function to rank the $n$ candidates, and returning the best one.
We refer to this embodiment optimizer as the \textbf{Zeroth-Order Optimizer}.

\textbf{Cross-Embodiment Control.}
We formulate a cross-embodiment controller that is embodiment-aware: conditioning on the embodiment~\cite{furuta2022system,schaff2019jointly,kurin2020my,gupta2022metamorph,lu2025bodygen,patel2025get,wang2019neural}, current state, and target motion, it predicts the expert actions that were used to track that motion.
Due to RoboToken's completeness, this conditioning provides critical information about the robot's kinematics, dynamics, and actuation capabilities to enable embodiment-aware action predictions.
Due to RoboToken's flexibility, heterogeneous state and action spaces (e.g., due to different DoF count) can be handled by varying the number of state and action tokens in the model's context.
When trained on data from diverse robots, the model can generalize to unseen embodiments at inference time.

\textbf{Model \& Training.}
For each token, we learn a token-type-specific linear projection from that type's vector space into the DiT's latent space, and add learned embeddings of the token's IDs.
Since our domain exhibits \(\mathrm{SE}(2)\) equivariance (not \(\mathrm{SE}(3)\), due to gravity), we apply planar transform augmentations.
Each token type's sequence is padded to the same maximum sequence length, then all token type sequences are concatenated together to form one long, multi-modal sequence.
We train the DiT using the DDIM scheduler~\cite{songdenoising}.
To speed up training, we subsample episode timesteps to form shorter, fixed-length sequences, and found that 8 timesteps were sufficient for the model to be effective for our use cases.
These timesteps are randomly sampled within an episode for the motion-to-robot task, and consecutively sampled for the control task to give it past action/future target information.
Hyperparameters are reported in Tab.~\ref{tab:model_hparams} and more model details in \S~\ref{sec:model_details}.

\vsqueeze{-4mm}
\subsection{Optimization with Dynamics Self-Guidance}
\label{sec:selfguidance}
\vsqueeze{-3mm}

In challenging optimization landscapes, zeroth-order optimizers may take many samples to return high value designs.
To address this, we introduce Dynamics Self-Guidance, a first-order diffusion guidance method that uses gradients from the model's noisy reward predictions to optimize its noisy embodiment predictions.
While prior works have explored converting dynamics into diffusion guidance signals, they require training a separate dynamics model~\cite{xu2024dynamics} or a differentiable simulator~\cite{wang2023diffusebot}.
Meanwhile, our approach uses a single, unified diffusion model as follows.

Given a differentiable reward, we can backpropagate its gradient to the embodiment tokens.
At step $k$, the DiT predicts $\epsilon_k$, which is used to denoise tokens from step $k$ to step $k-1$.
Thus, even if the full diffusion process is not differentiable, per-step reward gradients can still guide samples toward higher-reward candidates.
We incorporate this gradient information following classifier-guided DDIM~\cite{dhariwal2021diffusion}.
At inference time, we run $n$ parallel guided diffusion processes, then return the best design ranked by the model. We call this embodiment optimizer \textbf{Dynamics Self-Guidance}.
\vsqueeze{-3mm}
\subsection{Data Collection \& Generation}
\label{sec:data}
\vsqueeze{-2mm}

\textbf{Target Motion Dataset.}
We collected 76 motion trajectories using an extension of the UMI gripper~\cite{chi2024universal,choi2026wild,gupta2025umionair}, which is built on top of ARKit (Fig.~\ref{fig:teaser}, left). %
These target trajectories include single end-effector motions for diverse skills like tossing, screwing, opening drawers, and scrubbing.
We use 56 trajectories for training, and 20 trajectories for validation.
In addition, we use UMI's bimanual dishwashing demonstrations for the bimanual design space.

\textbf{Design Spaces.}
We picked three design spaces to stress-test three co-design challenges:
\begin{itemize}[leftmargin=3mm,itemsep=0pt]
    \vsqueeze{-2mm}
    \item \textbf{Kinematic Design}:
          The Trossen ViperX 300S design space (Fig.~\ref{fig:qualitative_hardware_opt_results}, right), varies mounting pose, degrees of freedom count, joint orientations, and link lengths.
          While fixed-base arms simplify control, they are unforgiving to poor kinematic design.
          Failure to reason about kinematic reachability will lead to large tracking penalties.
    \item \textbf{Dynamic Control}:
          The quadruped manipulator design space (Fig.~\ref{fig:qualitative_hardware_opt_results}, left) includes leg design variations (e.g., serial vs.\ spring-loaded linkage), arm mounts (fixed vs.\ sliding rail), on top of length variations for torso, legs, and arms.
          To successfully design high-value quadruped manipulators, the algorithm must reason through each design's whole-body control performance, paying attention to how stability and inertia trade off with agility for precise, dynamic tracking.
    \item \textbf{Task Complexity}: The mobile bimanual design space (Fig.~\ref{fig:teaser}) includes different spine designs (fixed, sliding, bending) on top of length variations.
          Bimanual, whole-body manipulation encapsulates tasks more complex than those involving only a single end-effector, including long-horizon, coordinated movements with many subtasks like in dishwashing. %
          \vsqueeze{-2mm}
\end{itemize}
Our design spaces are combinatorial, with discrete and continuous choices, providing diverse RoboTokens.
Next, we discuss how all generated robots are controlled to track target motions, completing our training dataset with state and action RoboTokens.

\textbf{DiffIK Control for Non-legged Robots.}
For the fixed-base and mobile bimanual design spaces above, we use Mink~\cite{Zakka_Mink_Python_inverse_2025}, a differential inverse kinematics library for MuJoCo~\cite{todorov2012mujoco}.
Since this kinematic solver doesn't consider dynamics, we account for joint position control lag with a fixed look-ahead time of 80\,ms.
We include terms for tracking, posture regularization, and damping.

\textbf{Whole-body Control for Legged Robots.}
Reinforcement learning (RL)~\cite{sutton1998reinforcement,schulman2017proximal} is the dominant paradigm for whole-body control of legged robots~\cite{ha2024umilegs,fu2023deep,ji2024exbody2,liu2024visual,portela2025whole,portela2024learning}.
However, training one policy per each robot~\cite{furuta2022system,patel2025get} is too costly.
Thus, we train one policy for each discrete variation (e.g., leg design), while including continuous variations (e.g., leg lengths) in the policy observation, allowing one policy to control many continuous variations.
We extend manipulation-centric WBCs~\cite{ha2024umilegs} to include continuous variation observations.
For each combination of 7 binary design choices, we train an RL policy used for data collection and design validation, yielding 128 RL experts in total.
We also investigated GPU-accelerated trajectory optimization~\cite{xue2025full}, but found that RL controller training was faster in total wall clock time for data generation.
We report more data generation details in \S~\ref{sec:data_details}.

\begin{figure*}[t]
    \centering
    \includegraphics[width=\linewidth]{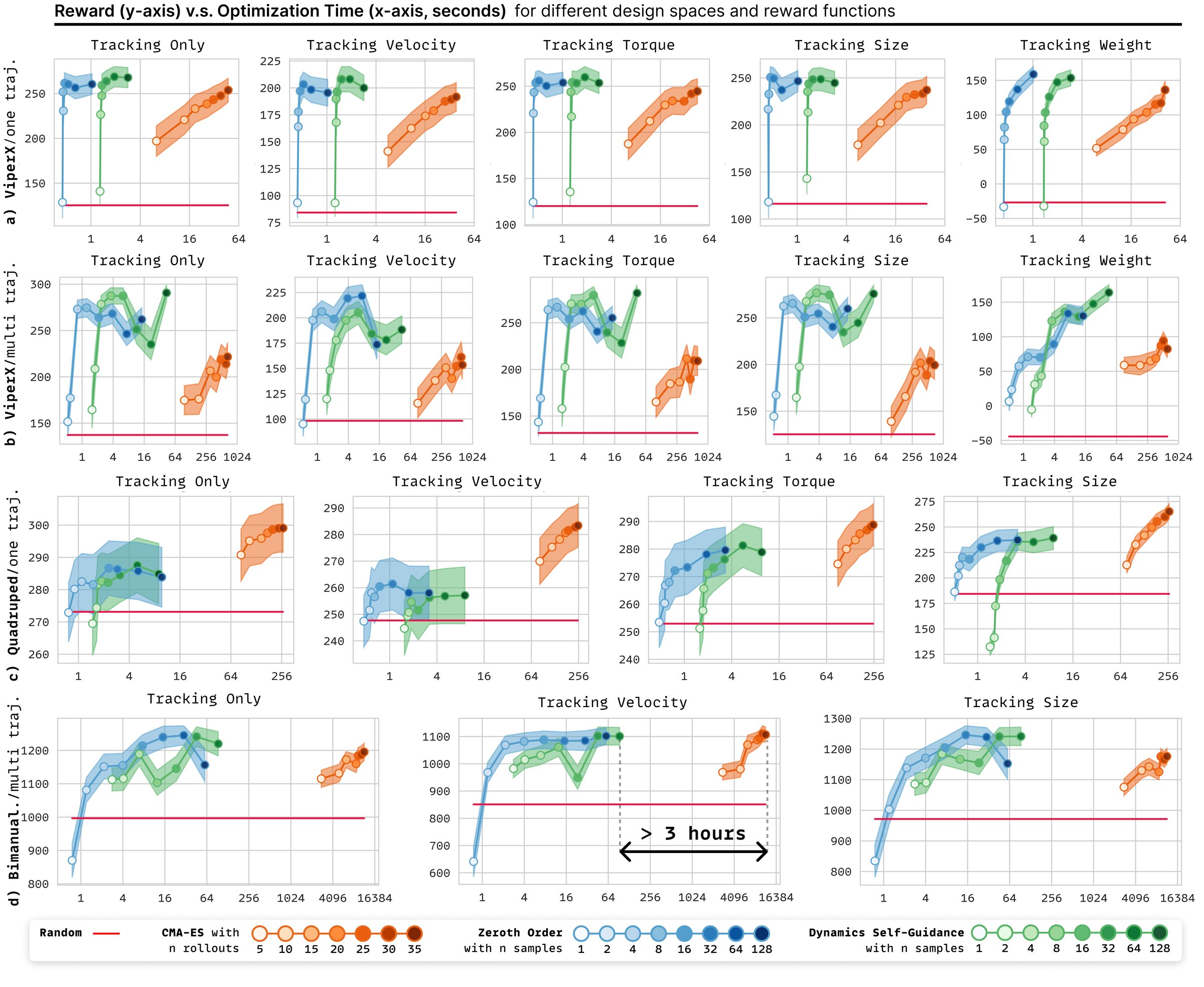}
    \caption{
        \small
        \textbf{Better Designs with Test-Time Scaling}.
        Across three design spaces and all rewards (\S~\ref{sec:additional_metrics}), both our \textbf{Zeroth Order} and \textbf{Dynamics Self-Guidance} tend to generate higher-reward designs when allowed more parallel samples.
        Our model also zero-shot optimizes multiple trajectories (b,d).%
        Our approach is significantly faster than \textbf{CMA-ES}, achieving similar or better performance while taking orders of magnitude less time.
    }
    \vsqueeze{-7mm}
    \label{fig:hardware_opt_results}
\end{figure*}

\section{Results}

\textbf{RoboToken Unifies Diverse Robot Embodiment Spaces.}
We tokenized eleven robots from MuJoCo Menagerie~\cite{menagerie2022github}, including a dexterous hand, fixed-base arms, quadrupeds (+ manipulators), bipeds, and humanoids.
Their masses span two orders of magnitude (Allegro V3's 0.65\,kg to Anymal C's 67.5\,kg), with DoF counts ranging from 6 active joints (UR5e) to 35 active/passive joints (Cassie), but all tokenize into RoboToken sequences of lengths 28 to 101.

\textbf{Generated Designs Lie on the Training Manifold.}
A natural question is what range of designs Transformer Transformer can produce. Empirically, our generator inherits properties theoretically expected of diffusion models~\cite{song2019generative,he2026demystifying}.
\emph{Manifold adherence}: since the score function is undefined off the training distribution, the learned denoiser steers samples toward the data manifold rather than beyond it. Generated designs interpolate within the training distribution but do not extrapolate beyond it---e.g., the model cannot produce hexapods from quadruped-only training, nor link lengths exceeding the training range. This is both a feature (physically plausible outputs by construction) and a limitation (extending the design space requires extending the dataset).
\emph{Mode coverage}: an unconditional model generates each robot type with probability matching the training distribution.
\emph{Cross-attribute coherence}: generated outputs respect the joint distribution of embodiment attributes---larger links carry proportional masses and inertias, motors are sized to the joints they actuate---ensuring physical feasibility without explicit constraints.

\textbf{Transformer Transformer Optimizes Embodiments for Unseen Reward Functions Zero-Shot.}
\label{sec:exp:hardware_opt}
At inference time, we optimize the generated robot hardware to maximize \emph{unseen} reward functions, conditioned on \emph{unseen} target tracking motions.
These rewards include terms for embodiment (e.g., size, weight) as well as behavior while tracking the target trajectories (e.g., tracking performance, torque/velocity usage).
We include reward definitions and additional metrics in the appendix (\S~\ref{sec:additional_metrics}).

\textbf{Approaches.}
(1) \textbf{Random} samples the discrete and continuous choices; (2) \textbf{CMA-ES}~\cite{hansen2001completely,hansen2019pycma}, a widely applied evolutionary algorithm in co-design~\cite{chen2020hardware,xu2021end,rajani2023co}, samples the discrete choices and then optimizes the continuous choices; (3) \textbf{Zeroth Order} samples $n$ robots from our model, ranks them with the model, and returns the highest ranked one
; (4) \textbf{Dynamics Self-Guidance} extends Zeroth Order by introducing gradient guidance.
To isolate embodiment optimization performance in our approaches, we train a model for each design space, and on only the motion-to-robot task.

\textbf{Evaluation Procedure}.
To account for optimization efficiency, we report wall-clock time taken for the optimization procedure to produce the optimized robot.
For each optimized robot, we validate the design in MuJoCo~\cite{todorov2012mujoco} with the design space's controller (RL experts/Mink \S~\ref{sec:data}), and report the average sum of rewards over the episode.
In Fig.~\ref{fig:hardware_opt_results}, we show the mean with 95\% confidence interval bars over held-out trajectories and CMA-ES/diffusion sampling 9 seeds.
All time measurements were conducted on a lightly loaded NVIDIA RTX 4090 and an Intel i9 processor core.

\textbf{Parallelized Design Evaluation.}
\label{para:exp:parallelized}
For the fixed-base and bimanual mobile design spaces (a,b,d), our model outperforms CMA-ES in terms of both speed and performance (Fig.~\ref{fig:hardware_opt_results}).
Beyond the advantages of GPU parallelization along the batch dimension of zeroth-order search, our approach enjoys a second axis of parallelization, along an episode's time dimension.
For CMA-ES, evaluating a candidate design requires running the controller and simulator over thousands of timesteps \emph{sequentially}.
In contrast, our model can reason about long-horizon dynamics (and thus performance) by considering a small set of episode timesteps \emph{in parallel}, enabled by our non-autoregressive formulation.
These two factors of parallelization multiply, allowing our approach to diffuse and evaluate 128 candidates much faster than CMA-ES can evaluate 5 candidates.

\begin{figure*}[t]
    \centering
    \includegraphics[width=\linewidth]{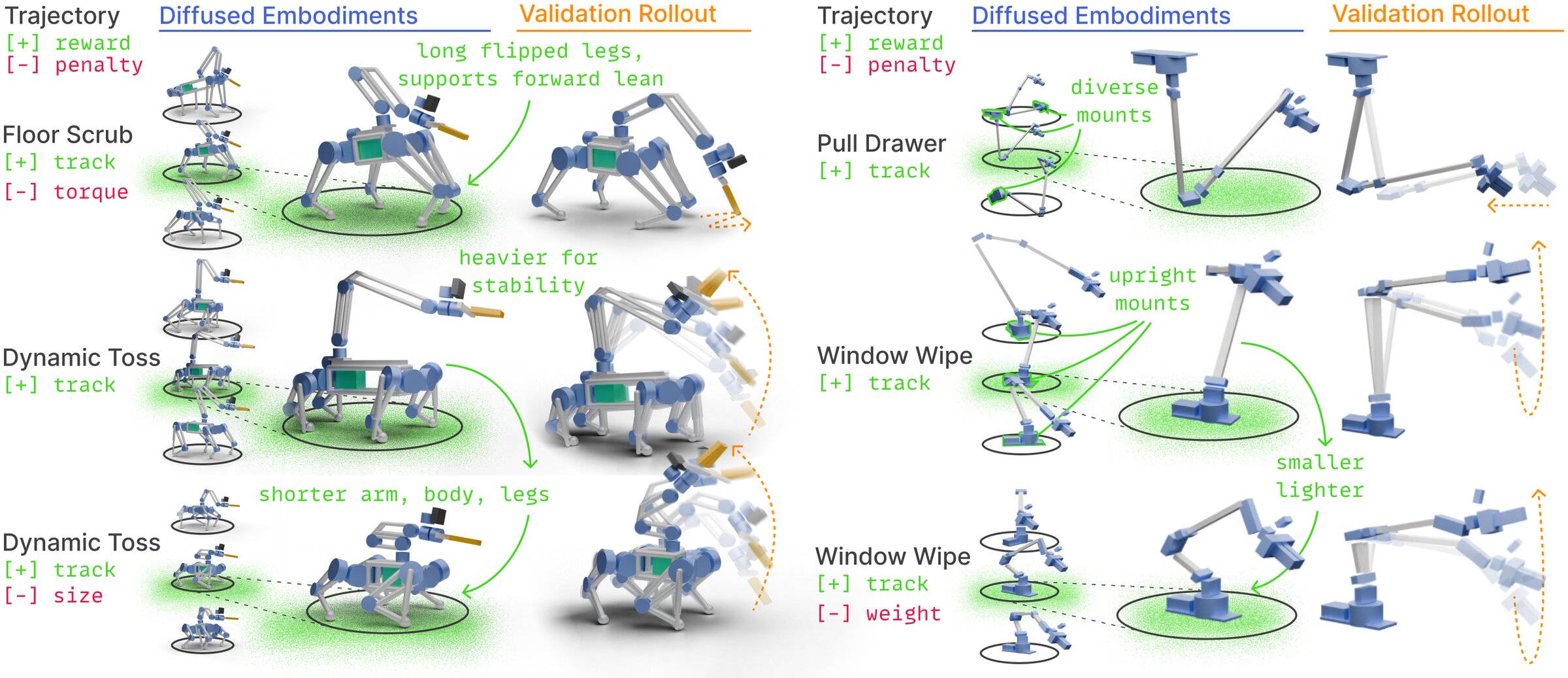}
    \caption{
        \small
        \textbf{Diffusion Generates Diverse Designs}.
        Our model diffuses embodiments with discrete (e.g., leg designs, degrees of freedom) and continuous (e.g., arm length, mounting points) variations that maximizes the specified reward functions.
        Changing the target trajectory (e.g., ``Floor Scrub'' $\rightarrow$ ``Dynamic Toss'') or the reward (e.g., ``size'' penalization) changes the optimization landscape, leading to different embodiment distributions (e.g., diverse $\rightarrow$ only upright mounts).
    }
    \vsqueeze{-7mm}
    \label{fig:qualitative_hardware_opt_results}
\end{figure*}
\textbf{Diffusion Generates Diverse Designs.}
To accurately track the tossing trajectory, our model generally prefers larger robots (more stable) with longer arms (faster velocity) (Fig.~\ref{fig:qualitative_hardware_opt_results}, middle left).
However, when the optimization landscape shifts to regularize the robot's size, our model converges on smaller robots.
Our model's ability to produce diverse robots for different reward functions arises from RoboToken's completeness,
which gives reward functions many handles to control the diffusion process. %
Meanwhile, the model's ability to produce diverse robots for the same reward function arises from using a diffusion formulation for Transformer Transformer.

\wrapclearance{14}
\begin{wrapfigure}{r}{0.5\textwidth}
    \centering
    \vsqueeze{-5mm}
    \includegraphics[width=\linewidth]{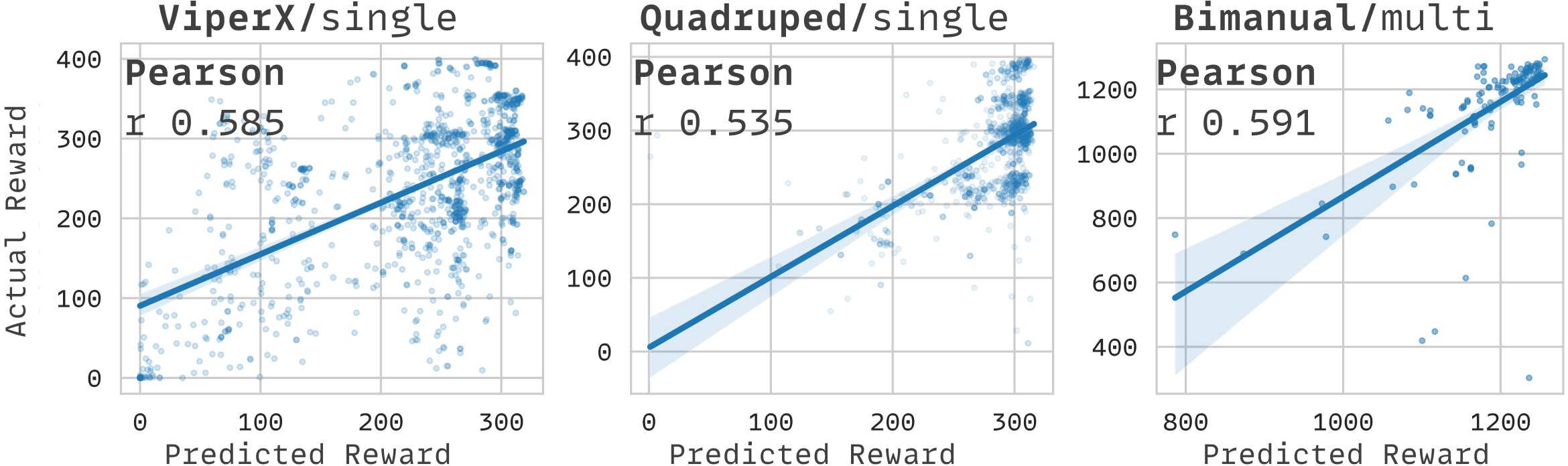}
    \vsqueeze{-3mm}
    \caption{
        \small
        \textbf{Predicted Reward Accuracy}.
        Learning legged robot dynamics means reasoning about how learned whole-body controllers navigate discontinuous contacts and termination risks from falling over.
        Thus, we observe higher correlation between predicted and actual rewards for ViperX and Bimanual than Quadruped design spaces.
    }
    \vsqueeze{-4mm}
    \label{fig:hallucination}
\end{wrapfigure}

\textbf{Diffusion Composition for Multi-Trajectory Optimization.}
We apply diffusion composition~\cite{liu2022compositional,du2023reduce} to enable zero-shot multi-trajectory optimization from only single trajectory training.
Running the composed diffusion process generates designs that perform favorably over many unseen trajectories simultaneously (20/26 trajs in Fig.~\ref{fig:hardware_opt_results}b,d).
Our approach parallelizes favorably, while CMA-ES's optimization time scales linearly with the number of trajectories, taking \emph{3 hours longer}.
Unlike many computational design prior works which focus on task-specific designs~\cite{kodnongbua2023computational,ha2021fit2form,xu2024dynamics,xu2021end,li2022learning,liu2024paperbot,allen2022physical}, our ability to optimize for multiple trajectories broadens its potential applications, such as in optimizing a generalist manipulation platform by conditioning on a motions from different tasks from a manipulation dataset~\cite{khazatsky2024droid}.

\textbf{Better Designs with Test-Time Compute.}
When given more test-time compute, language reasoning models generate better outputs~\cite{openai_learning_to_reason_2024,guo2025deepseek,comanici2025gemini}.
We observe a similar phenomenon in our model (Fig.~\ref{fig:hardware_opt_results}), where increasing test-time compute by searching over more parallel seeds~\cite{brown2024large} allows the model to produce better designs.
Notably, this behavior holds across all design spaces and reward functions, emerging purely from learning to model robot embodiments and dynamics.
Although our model can refine its embodiment design for up to a minute (Fig.~\ref{fig:hardware_opt_results}d, guided), performance does not robustly increase with test-time compute, as observed in earlier language models~\cite{brown2024large,ballon2025relationship,wu2024inference}.
Future work could investigate improving dynamics (and thus reward) modeling accuracy (Fig.~\ref{fig:hallucination}) and alternative inference schemes~\cite{brown2024large,wu2024inference} to improve test-time compute scaling.

\begin{wrapfigure}{r}{0.35\textwidth}
    \centering
    \vsqueeze{-5mm}
    \includegraphics[width=\linewidth]{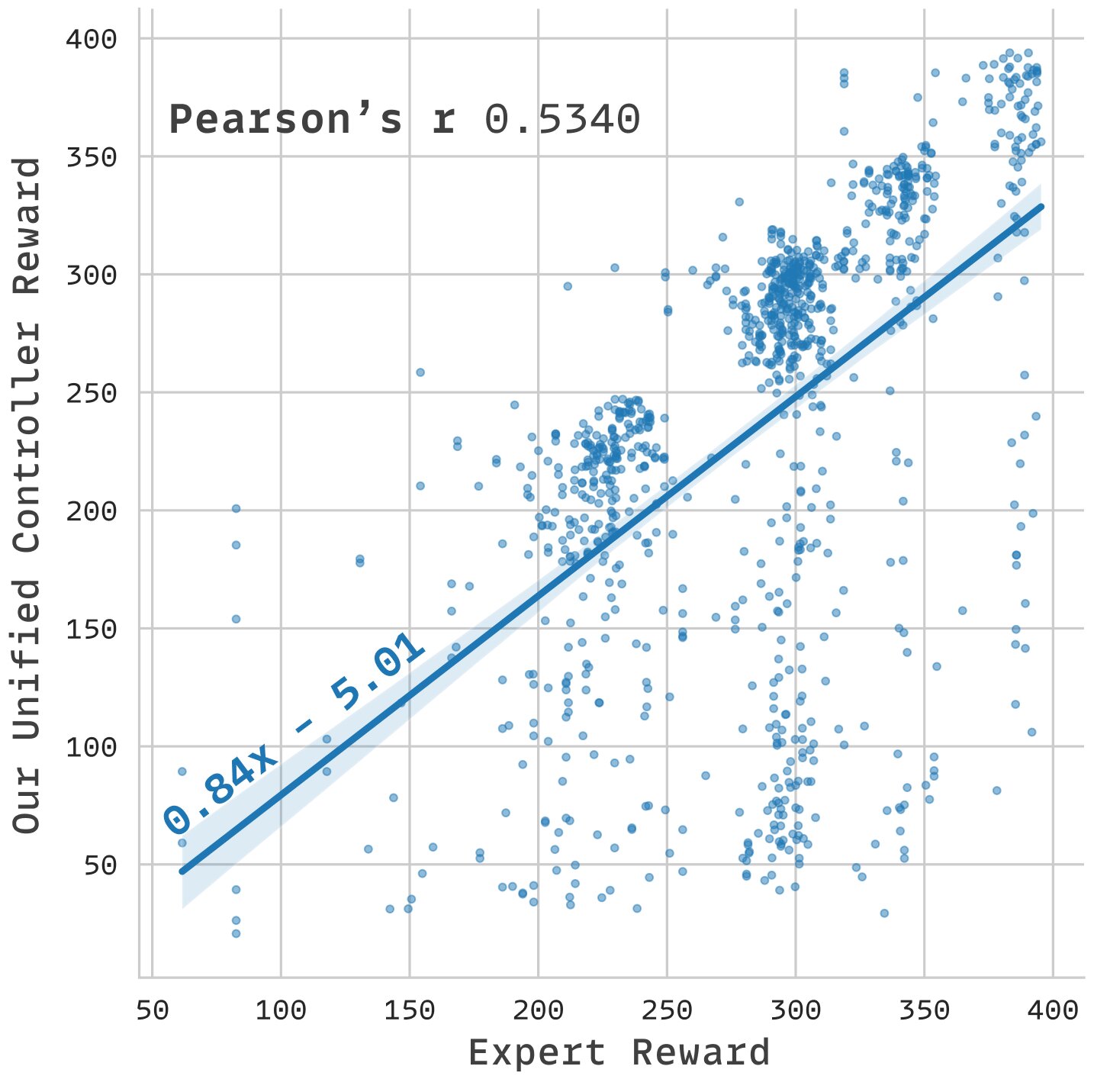}
    \vsqueeze{-7mm}
    \caption{
        \small
        \textbf{Whole-body Control of Generated Quadrupeds}.
        After generating optimized robots, the same model can directly control its designs.
    }
    \vsqueeze{-5mm}
    \label{fig:control_selfvalidation}
\end{wrapfigure}
\textbf{Guidance Substitutes for Search.}
Dynamics Self-Guidance improves each individual sample, whereas Zeroth Order improves by selecting the best of many, making the two substitutes rather than complements.
Their gap is therefore largest where search has least to work with: with a single sample, Dynamics Self-Guidance leads Zeroth Order by 242 to 341 reward on the multi-trajectory bimanual rows, where one design must serve all 26 trajectories.
Given a large enough sampling budget, Zeroth Order's selection closes this gap and the two become indistinguishable (\S~\ref{sec:additional_metrics}).
Guidance is thus most valuable when samples are expensive---larger models, longer episodes, or interactive use---rather than when they are cheap enough to brute force.

\textbf{Transformer Transformer Controls Unseen Robots.}
We study our model's ability to \emph{faithfully} self-validate designs through controlling its generated embodiments.
Here, faithful means embodiments should achieve similar reward between experts and self-validation.
To perform well, the learned controller must transfer to diffused embodiments despite being trained on clean, procedurally generated robots.
Failure to do so may lead to the robot falling over, truncating the episode and the sum of rewards.
Despite this challenge, we observe a positive correlation (Pearson's r: 0.53, Fig.~\ref{fig:control_selfvalidation}).
Although there are some outlier embodiments, most cluster along the diagonal, demonstrating promising first steps towards replacing the 128 RL experts with a single, unified cross-embodiment controller.
We report more results in the appendix (Fig.~\ref{fig:bimanual_control}).

\wrapclearance{20}
\begin{wrapfigure}{r}{0.47\textwidth}
    \vsqueeze{-5mm}
    \centering
    \includegraphics[width=\linewidth]{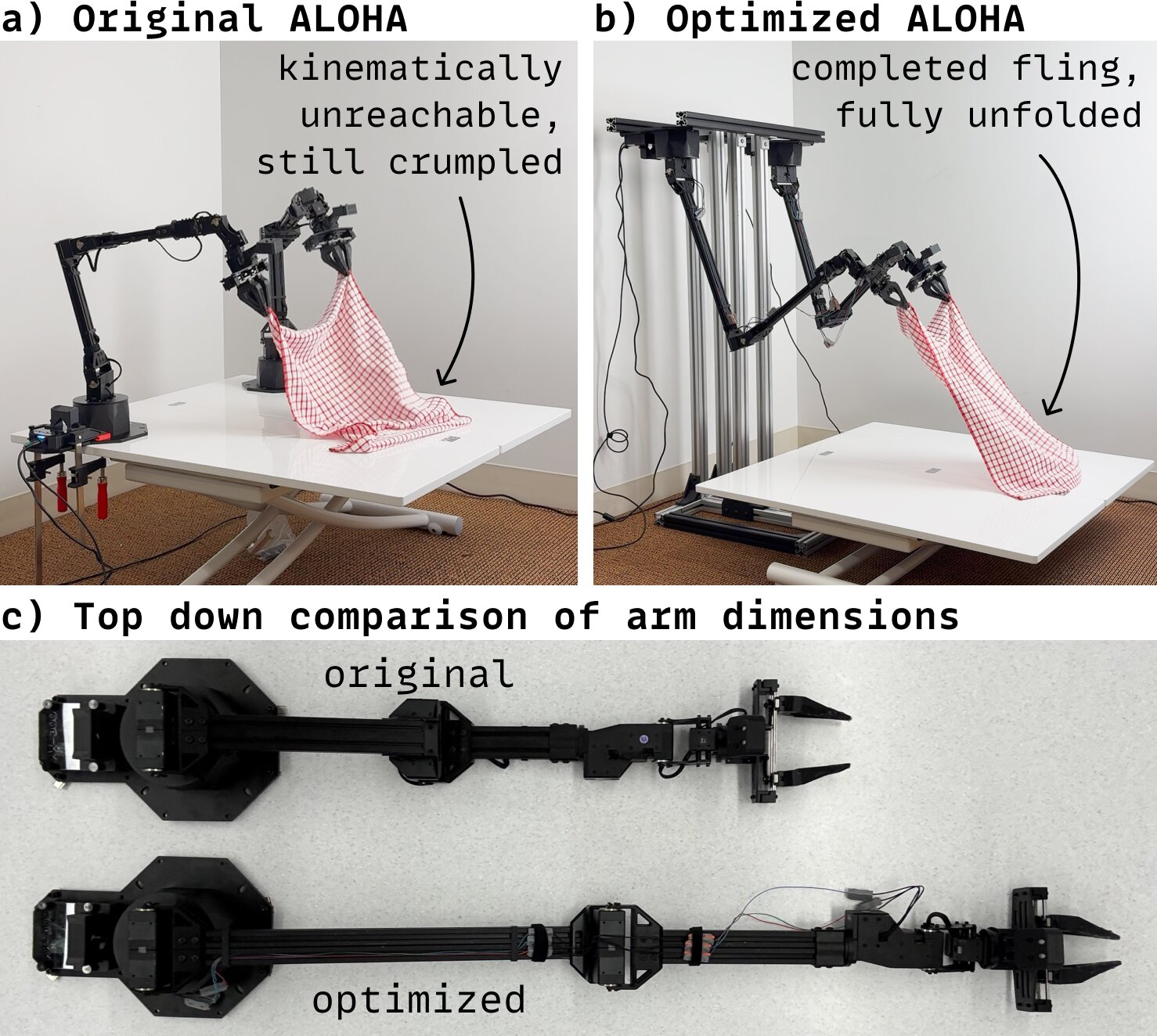}
    \caption{
        \small
        \textbf{Dynamic Cloth Unfolding.}
        To optimize ALOHA for high speed flings, our model found better mounting point and link dimensions, reducing tracking error by $73\%$ and max joint speed by $30\%$.
    }
    \vspace{-5mm}%
    \ifsqueeze\else\vspace{-\baselineskip}\fi%
    \label{fig:real_world}
\end{wrapfigure}

\begin{figure}[t]
      \centering
      \includegraphics[width=\linewidth]{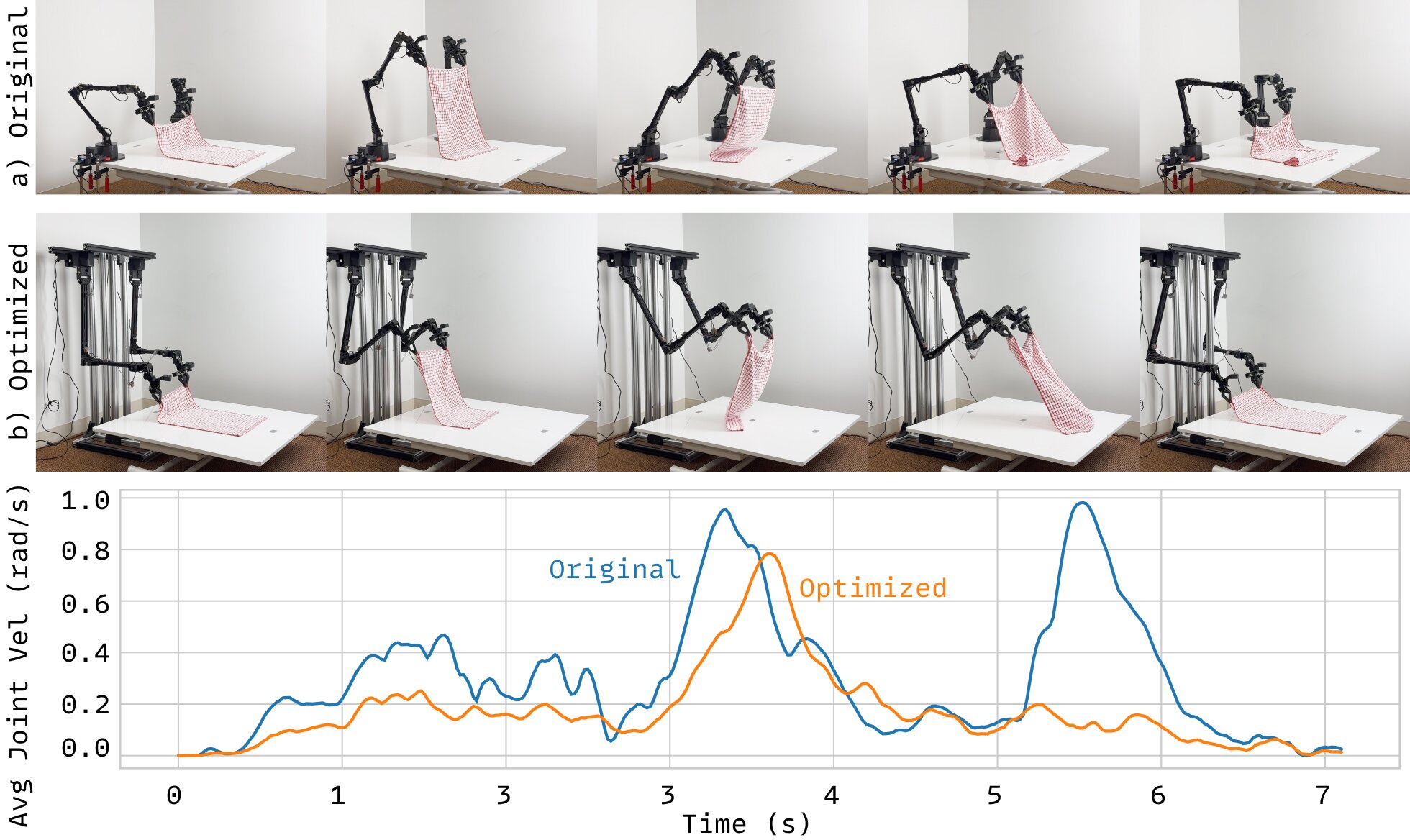}
      \caption{
            \small
            \textbf{Real World Unfolding Sequence}. Compared to the original ALOHA design (a), we observe a lower max joint speed and less peaks in our model's optimized design (b).
      }
      \label{fig:real_world_velocity}
\end{figure}

\textbf{Real-world design validation.}
We fabricated an ALOHA setup~\cite{zhao2023learning,zhao2024aloha} optimized by our model for the ``Tracking Velocity'' reward (Fig.~\ref{fig:real_world},\ref{fig:real_world_velocity}).
We focus on flinging for cloth unfolding~\cite{ha2022flingbot}, which stress-tests the framework with challenging dynamic manipulation and demands robustness to modeling errors and unmodeled disturbances (e.g., aerodynamic drag, cloth friction and weight).
Compared to the original design, the optimized design achieved a 30\% reduction in maximum joint speed ($2.57 \rightarrow 1.82$ rad/s) and a 73\% reduction in tracking position error ($13.0 \rightarrow 3.5$ cm).
The optimized design featured link lengths long enough to track the full motion while remaining light enough to be supported by ALOHA's Dynamixels motors.
It also mounted the arms upside down behind the workspace, enabling a more efficient underarm swing instead of an overhead fling.

\vsqueeze{-3mm}
\section{Related Works}
\vsqueeze{-3mm}

\textbf{Embodiment Design and Optimization.}
Traditional gradient-based methods require limited embodiment representations (e.g., cage-based~\cite{xu2021end,li2022learning}), heuristics~\cite{kodnongbua2023computational,ha2018computational}, and don't extend to domains with complex control and contacts~\cite{suh2022pathologies,xu2022accelerated,ha2018computational,ha2018cooptimization} (e.g., RL for legged robots).
Data-driven methods are flexible but overfit to one reward~\cite{ha2021fit2form,liu2024paperbot}.
Dynamics-guided generation~\cite{xu2024dynamics,allen2022physical} generalizes across rewards with a dynamics model whose gradients steer design, but still opts for fixed open-loop~\cite{ha2021fit2form,xu2024dynamics} or simplified kinematic actions~\cite{kodnongbua2023computational,kulz2025design}, sidestepping the full control problem.
Transformer Transformer instead learns a dynamics model over any action space for design.

\textbf{Cross-Embodiment Control.}
GPU-based simulators~\cite{rudin2022learning,Genesis} have made RL~\cite{sutton1998reinforcement,schulman2017proximal} the dominant approach for locomotion and whole-body control~\cite{kumar2021rma,rudin2022learning,zhuang2023robot,fu2021minimizing,margolis2023walk,ha2024umilegs,fu2023deep,ji2024exbody2,liu2024visual,portela2025whole,portela2024learning}, with cross-embodiment extensions by distilling single-embodiment experts into a shared policy~\cite{furuta2022system,patel2025get} that conditions on the embodiment~\cite{furuta2022system,schaff2019jointly,kurin2020my,gupta2022metamorph,lu2025bodygen,patel2025get,wang2019neural}.
However, their embodiment representations are typically incomplete.

\textbf{Limitations}
Towards a more complete co-design formulation for manipulation, RoboToken's scope should be expanded to support complex geometry~\cite{ha2021fit2form,xu2024dynamics,wang2023diffusebot,xu2021end,li2022learning}, scene/object~\cite{mandireal2code}, and tactile information~\cite{choi2026wild}.
Further, scaling RoboToken data generation (e.g., more efficient legged control) to more diverse robots will be critical towards a foundation model that reasons about the robot's embodiment as first-class tokens.
Ultimately, such advances would allow us to explore robot morphologies as diverse as the manipulation tasks they perform, moving beyond robot hardware as a static constraint.

\textbf{Robot Co-design.}
Evolutionary algorithms~\cite{sims2023evolving,lipson2000automatic,de2017evolutionary,langton1997artificial,cheney2014unshackling} showed that, given a simulator, enough compute will find high-value designs.
Subsequent data-driven work accelerates this loop with learned design generators~\cite{yuan2021transform2act,ha2019reinforcement,schaff2019jointly,hu2023glso,hu2022modular,wang2023diffusebot,lu2025bodygen}, critics~\cite{zhao2020robogrammar,xu2021multi,hu2022modular,fay2025cross}, or controllers~\cite{schaff2019jointly,hu2022modular,yuan2021transform2act,fay2025cross,chen2020hardware,lu2025bodygen,wang2019neural}, but these components are trained on disjoint representations and objectives, with iterative pipelines that are hard to speed up.
Transformer Transformer instead unifies generator, critic, and controller into one model, consolidating the co-design process into a single GPU-accelerated diffusion process.

\section{Conclusion}

We proposed Transformer Transformer, a DiT that performs generative robot design and control, enabling a one-stop shop for motion-conditioned robot co-design.
Our simulation results across three designs spanning fixed-base, legged, and bimanual mobile manipulation demonstrated the model's efficiency, performance, and zero-shot capabilities to unseen rewards, and our real world results validated the model's robustness to misspecified/unmodelled dynamics and practical usage.

\section{Acknowledgment}
This work was supported in part by the
NSF Award \#2143601, \#2037101, \#2132519, \#2153854, Google and Stanford HAI. The views and conclusions contained herein are those of the authors and should not be interpreted as necessarily representing the official policies, either expressed or implied, of the sponsors.

\ifsqueeze\newpage\fi %

\bibliographystyle{corlabbrvnat}
\bibliography{references}

\newpage

\section{Additional Experiments}

\begin{maybewrap}{r}{0.5\textwidth}
      \vsqueeze{-26mm}
      \centering
      \includegraphics[width=\linewidth]{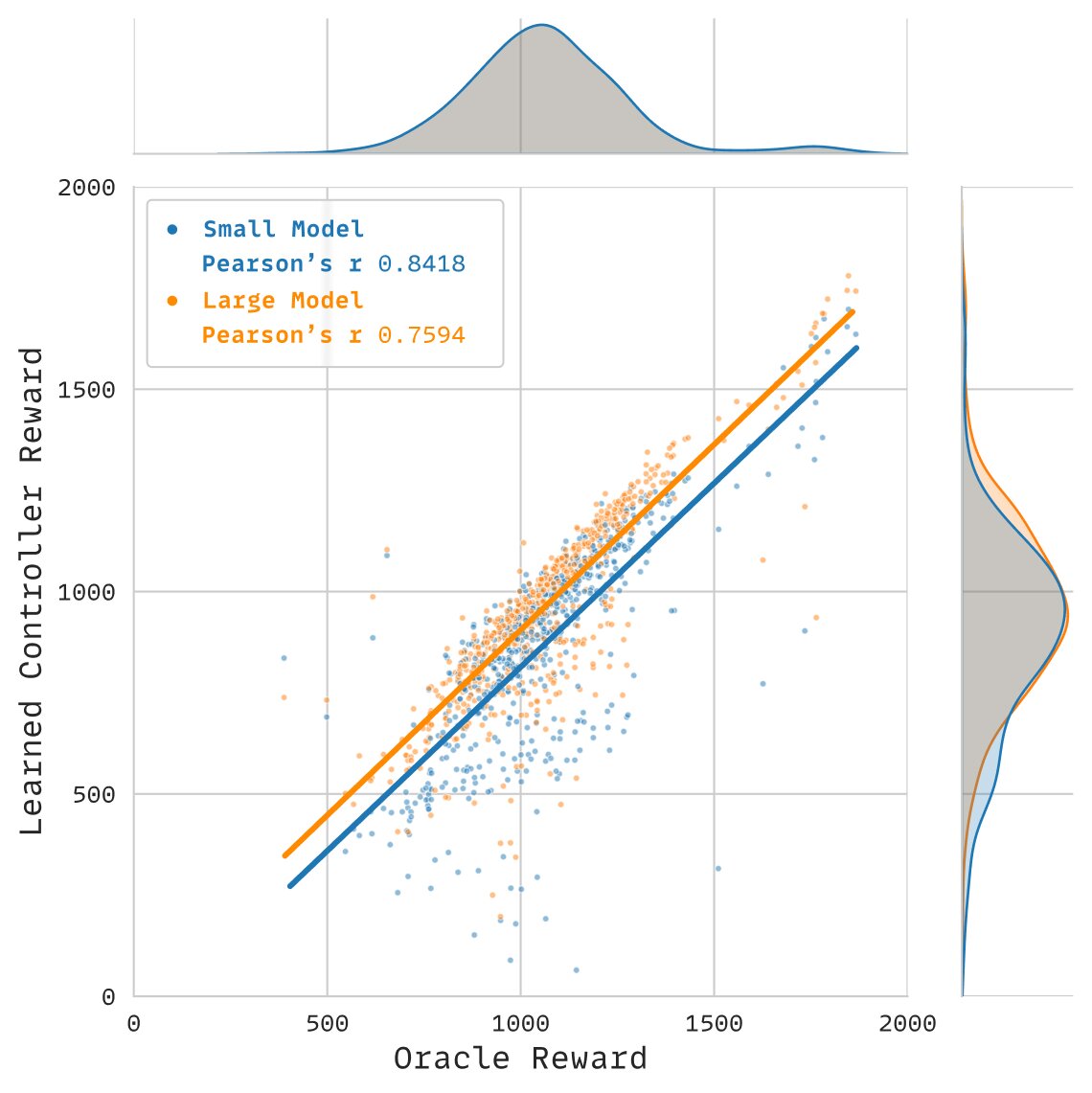}
      \vsqueeze{-6mm}
      \caption{
            \textbf{Bimanual Mobile Control Validation}.
            For each embodiment, we control it using our cross-embodiment controllers (small/large) and the oracle Mink~\cite{Zakka_Mink_Python_inverse_2025} controller.
            Our controllers have strong correlation with the oracle's performance.
            In other words, using our cross-embodiment controllers to evaluate robot designs is faithful to the oracle robot evaluation.
      }
      \vsqueeze{-18mm}
      \label{fig:bimanual_control}
\end{maybewrap}

In the context of robot co-design, a learned cross-embodiment controller is practically useful only for validating robot embodiments that benefit from costly learned controllers, such as legged robots.
However, robots with fast (oracle) kinematic controllers provide faster data generation and thus represent a practical design space to study cross-embodiment control.
In this section, we include capacity ablations when training a unified model, with cross-embodiment control (\S~\ref{sec:bimanual_control}) and motion-to-robot  (\S~\ref{sec:bimanual_robot_design}) experiments.

\begin{maybewrap}{r}{0.4\textwidth}
      \centering
      \includegraphics[width=\linewidth]{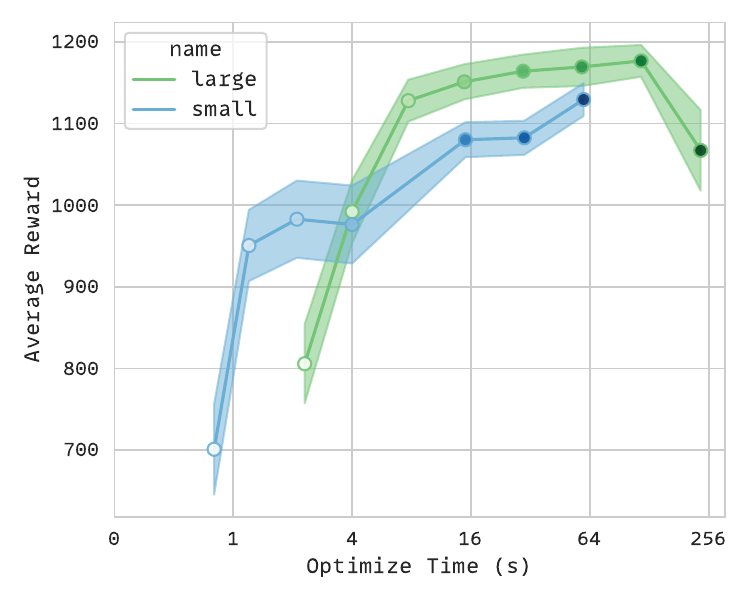}
      \vsqueeze{-8mm}
      \caption{
            \small
            \textbf{Model Capacity Ablation}. Although larger models perform better, their inference time also takes longer.
      }
      \vsqueeze{-14mm}
      \label{fig:hardware_opt_model_capacity}
\end{maybewrap}

\subsection{Bimanual Mobile Cross-Embodiment Control}
\label{sec:bimanual_control}
\textbf{Setup.}
We train two model sizes, a small model with 11.6M parameters and a large model with 63.6M parameters, on both the motion-to-robot task and the cross-embodiment controller task.
We evaluate the performance for the controller on procedurally generated robots in the mobile bimanual design space using the 26 validation trajectories from the UMI~\cite{chi2024universal} bimanual dish washing dataset.
For each trajectory, we include 5 continuous variations per discrete variation of the robot, yielding 25 robots over 650 episodes in total.
Since these robots are randomly generated, some embodiments with suboptimal designs exhibit poor control performance for all control methods (hence the higher tracking error).

\textbf{Results.}
We report the tracking reward defined in~\eqref{eq:tracking_only}, along with other tracking metrics, in Table~\ref{tab:bimanual_control}.
In Fig.~\ref{fig:bimanual_control}, we observe a strong positive correlation between the tracking performance of our controllers and the oracle Mink controller.
We normalize for the number of gradient steps, and report that the larger model is closer to oracle performance in terms of both tracking metrics and Pearson's correlation.
However, when normalizing for training FLOPs, the smaller model should receive $5.5\times$ as many training steps.
We leave this investigation, along with other model distillation techniques, for future work.

\subsection{Bimanual Mobile Motion-to-Robot Optimization}
\label{sec:bimanual_robot_design}

\textbf{Setup.}
We use the unified models with 11.6M and 63.6M parameters from the previous section (\S~\ref{sec:bimanual_control}) and apply the Zeroth Order optimizer for the multi-trajectory optimization setting.
As in the main paper, each robot is optimized for all 26 validation trajectories in the UMI dish washing dataset simultaneously, and we repeat for 9 inference seeds.
Unlike in the main paper, we use our learned controller (\S~\ref{sec:bimanual_control}) instead of the oracle controller.
We focus these results on the ``Tracking Only'' reward.

\textbf{Results.}
From Fig.~\ref{fig:hardware_opt_model_capacity}, we observe that the large model performs better.
However, for the same number of parallel samples, the large model requires $4\times$ longer runtime.

\begin{table}[t]
      \centering
      \begin{tabular}{lrrrr}
            \toprule
            Approach      & Pos Err             & Orn Err              & Survival          & Reward     \\
                          & \it{cm}$\downarrow$ & \it{deg}$\downarrow$ & \it{\%}$\uparrow$ & $\uparrow$ \\
            \midrule
            Ours (Small)  & 6.6                 & 10.7                 & 98.9              & 884.9      \\
            Ours (Large)  & 5.9                 & 9.6                  & 99.5              & 957.0      \\
            Mink (Oracle) & 4.8                 & 7.6                  & 100.0             & 1064.3     \\
            \bottomrule
      \end{tabular}
      \caption{\textbf{Controller Performance against Oracle}}
      \label{tab:bimanual_control}

\end{table}
\section{Additional Metrics from Main Experiments}
\label{sec:additional_metrics}

This section defines all reward terms used for the experiments and reports reward-specific metrics for the experiments presented in Figure 5 in the main paper.

\subsection{Tracking Only}
\label{sec:appendix:tracking_only}
\textbf{Reward Definition.}
For a given target pose trajectory $\{p^t_{target},o^t_{target}\}_{t \in T}$ and an achieved robot end-effector pose trajectory $\{p^t_{achieved},o^t_{achieved}\}_{t \in T}$ each of length $T$, we compute the position error $\epsilon^t_{p}$ as $||p^t_{target}-p^t_{achieved}||_2$ and the orientation error $\epsilon^t_{o}$ as $\arccos\left(\frac{\Tr(R) - 1}{2}\right)$ where $R$ is the rotation from $o^t_{achieved}$ to $o^t_{target}$.
Based on these tracking errors per time step, the tracking reward for an episode of length $T$ is:
\begin{equation}
      \sum_{t}^T \exp\left(-\frac{{||\epsilon^t_{p}||}^2}{\sigma_p} - \frac{{||\epsilon^t_{o}||}^2}{\sigma_o}\right)
      \label{eq:tracking_only}
\end{equation}
Intuitively, the reward is 1 if both position and orientation errors are zero, and less than 1 if either error is non-zero.
$\sigma_p$ and $\sigma_o$ determine the reward falloff as tracking error increases and are set to $0.01$ and $0.5$, respectively, for all experiments.
Exponentiating the negative tracking error is a common reward function form used in reinforcement learning for robotics~\cite{rudin2022learning,margolis2023walk,ha2024umilegs}.
An episode is terminated if the robot falls over (in the case of the quadruped) or if $\epsilon^t_{p}$ exceeds a maximum allowed position error threshold.
This threshold is set to 50cm for the ViperX and mobile bimanual design spaces, and 80cm for the quadruped design space.
For the bimanual case, we define $\epsilon^t_{p}$ and $\epsilon^t_{o}$ to be the sum of tracking position and orientation errors over both end-effectors.
This means high reward is only possible if position and orientation errors for both end-effectors are low.

\textbf{Reward-Specific Metrics.}
We report position and orientation tracking error averaged over all time steps and all end-effectors, the optimization time in seconds, and the survival rate (proportion of episodes that did not terminate early).
For each approach, we use the value of $n$ that achieves the highest reward.
For CMA-ES, $n$ is the maximum number of rollouts per trajectory.
For Zeroth Order and Dynamics Self-Guidance (DGS), $n$ is the number of samples.

\begin{table}[H]
      \centering
      \begin{tabular}{lrrrr}
            \toprule
            Approach & Pos Err             & Orn Err              & Survival          & Optimize Time        \\
                     & \it{cm}$\downarrow$ & \it{deg}$\downarrow$ & \it{\%}$\uparrow$ & \it{sec}$\downarrow$ \\
            \midrule
            Random   & 19.4                & 8.9                  & 72.2              & -                    \\ %
            CMA-ES   & 5.0                 & 5.7                  & 98.3              & 47.5                 \\ %
            Zeroth   & \textbf{4.1}        & 4.2                  & 96.1              & \textbf{0.5}         \\ %
            DGS      & \textbf{4.1}        & \textbf{3.9}         & \textbf{99.4}     & 2.8                  \\ %
            \bottomrule
      \end{tabular}
      \caption{\textbf{ViperX, Single Trajectory}}
      \label{tab:viperx:tracking_only}
\end{table}
\begin{table}[H]
      \centering
      \begin{tabular}{lrrrr}
            \toprule
            Approach & Pos Err             & Orn Err              & Survival          & Optimize Time        \\
                     & \it{cm}$\downarrow$ & \it{deg}$\downarrow$ & \it{\%}$\uparrow$ & \it{sec}$\downarrow$ \\
            \midrule
            Random   & 17.0                & 9.4                  & 80.6              & -                    \\ %
            CMA-ES   & 7.1                 & 6.1                  & 86.7              & 663.8                \\ %
            Zeroth   & 3.4                 & 3.9                  & 95.6              & \textbf{1.2}         \\ %
            DGS      & \textbf{2.4}        & \textbf{3.5}         & \textbf{98.9}     & 43.5                 \\ %
            \bottomrule
      \end{tabular}
      \caption{\textbf{ViperX, Multi-Trajectory}}
\end{table}
\begin{table}[H]
      \centering
      \begin{tabular}{lrrrr}
            \toprule
            Approach & Pos Err             & Orn Err              & Survival          & Optimize Time        \\
                     & \it{cm}$\downarrow$ & \it{deg}$\downarrow$ & \it{\%}$\uparrow$ & \it{sec}$\downarrow$ \\
            \midrule
            Random   & 3.4                 & 8.4                  & 96.1              & -                    \\ %
            CMA-ES   & \textbf{1.9}        & \textbf{5.8}         & \textbf{100.0}    & 265.5                \\ %
            Zeroth   & 2.6                 & 7.4                  & 99.4              & 0.8                  \\ %
            DGS      & 2.6                 & 7.4                  & 99.4              & 5.0                  \\ %
            \bottomrule
      \end{tabular}
      \caption{\textbf{Quadruped, Single-Trajectory}}
\end{table}
\begin{table}[H]
      \centering
      \begin{tabular}{lrrrr}
            \toprule
            Approach & Pos Err             & Orn Err              & Survival          & Optimize Time        \\
                     & \it{cm}$\downarrow$ & \it{deg}$\downarrow$ & \it{\%}$\uparrow$ & \it{sec}$\downarrow$ \\
            \midrule
            Random   & 2.8                 & 4.4                  & 99.6              & -                    \\ %
            CMA-ES   & \textbf{1.7}        & \textbf{2.5}         & \textbf{100.0}    & 11505.2              \\ %
            Zeroth   & 1.8                 & \textbf{2.5}         & 99.6              & \textbf{20.7}        \\ %
            DGS      & \textbf{1.7}        & 2.7                  & 99.1              & 30.8                 \\ %
            \bottomrule
      \end{tabular}
      \caption{\textbf{Bimanual, Multi-Trajectory}}
\end{table}

\subsection{Tracking Torque}

\textbf{Reward Definition.}
Let $\tau^t$ be the total torque in all of the robot's motors at timestep $t$.
The tracking torque reward is:
\begin{equation}
      \sum_{t}^T \max\left(\exp\left(-\frac{{||\epsilon^t_{p}||}^2}{\sigma_p} - \frac{{||\epsilon^t_{o}||}^2}{\sigma_o}\right) -  \alpha_{torque}||\tau^t||^2,0\right)
      \label{eq:tracking_torque}
\end{equation}
where the torque weight $\alpha_{torque}$ is set to $5 \times 10^{-5}$ in all experiments.
If the torque is too large, it will outweigh the non-negative tracking term.
Avoiding negative rewards is common practice~\cite{rudin2022learning,margolis2023walk,ha2024umilegs}, preventing scenarios where large negative rewards encourage agents to terminate episodes early.

\textbf{Reward-Specific Metrics.}
In addition to the tracking metrics (\S~\ref{sec:appendix:tracking_only}), we also report averaged $||\tau^t||_2$.

\begin{table}[H]
      \centering
      \begin{tabular}{lrrrrr}
            \toprule
            Approach & Pos Err             & Orn Err              & Survival          & Torque              & Optimize Time        \\
                     & \it{cm}$\downarrow$ & \it{deg}$\downarrow$ & \it{\%}$\uparrow$ & \it{Nm}$\downarrow$ & \it{sec}$\downarrow$ \\
            \midrule
            Random   & 19.4                & 8.9                  & 72.2              & \textbf{3.8}        & -                    \\ %
            CMA-ES   & 5.9                 & 5.7                  & 95.0              & 4.1                 & 45.4                 \\ %
            Zeroth   & \textbf{4.1}        & 4.5                  & 97.2              & 4.4                 & \textbf{0.5}         \\ %
            DGS      & \textbf{4.1}        & \textbf{4.1}         & \textbf{98.3}     & 4.5                 & 1.9                  \\
            \bottomrule
      \end{tabular}
      \caption{\textbf{ViperX, Single Trajectory}}
\end{table}

\begin{table}[H]
      \centering
      \begin{tabular}{lrrrrr}
            \toprule
            Approach & Pos Err             & Orn Err              & Survival          & Torque              & Optimize Time        \\
                     & \it{cm}$\downarrow$ & \it{deg}$\downarrow$ & \it{\%}$\uparrow$ & \it{Nm}$\downarrow$ & \it{sec}$\downarrow$ \\
            \midrule
            Random   & 17.0                & 9.4                  & 80.6              & \textbf{4.0}        & -                    \\ %
            CMA-ES   & 7.5                 & 5.9                  & 89.4              & 4.4                 & 644.4                \\ %
            Zeroth   & 3.4                 & 3.9                  & 95.6              & 4.6                 & \textbf{1.2}         \\ %
            DGS      & \textbf{2.5}        & \textbf{3.5}         & \textbf{98.9}     & 4.6                 & 43.4                 \\ %
            \bottomrule
      \end{tabular}
      \caption{\textbf{ViperX, Multi Trajectory}}
\end{table}
\begin{table}[H]
      \centering
      \begin{tabular}{lrrrrr}
            \toprule
            Approach & Pos Err             & Orn Err              & Survival          & Torque              & Optimize Time        \\
                     & \it{cm}$\downarrow$ & \it{deg}$\downarrow$ & \it{\%}$\uparrow$ & \it{Nm}$\downarrow$ & \it{sec}$\downarrow$ \\
            \midrule
            Random   & 3.4                 & 8.4                  & 96.1              & 4.6                 & -                    \\
            CMA-ES   & \textbf{2.0}        & \textbf{6.0}         & \textbf{100.0}    & 3.2                 & 250.0                \\
            Zeroth   & 2.5                 & 7.1                  & 99.4              & \textbf{3.1}        & \textbf{3.3}         \\ %
            DGS      & 2.5                 & 7.0                  & 98.9              & 3.2                 & 5.5                  \\
            \bottomrule
      \end{tabular}
      \caption{\textbf{Quadruped, Single Trajectory}}
\end{table}

\subsection{Tracking Velocity}

\textbf{Reward Definition.}
This reward follows \eqref{eq:tracking_torque}, replacing $\alpha_{torque}||\tau^t||^2$ with $\alpha_{velocity}||\dot{q}^t||^2$, where $\dot{q}^t$ is each actuator's velocity at time step $t$.
$\alpha_{velocity}$ is set to 0.1 for the ViperX, 0.5 for the mobile bimanual design space, and 0.005 for the quadruped design space.

\textbf{Reward-Specific Metrics.}
In addition to the tracking metrics (\S~\ref{sec:appendix:tracking_only}), we also report averaged $||\dot{q}^t||_2$.

\begin{table}[H]
      \centering
      \begin{tabular}{lrrrrr}
            \toprule
            Approach & Pos Err             & Orn Err              & Survival          & Velocity               & Optimize Time        \\
                     & \it{cm}$\downarrow$ & \it{deg}$\downarrow$ & \it{\%}$\uparrow$ & \it{rad/s}$\downarrow$ & \it{sec}$\downarrow$ \\
            \midrule
            Random   & 19.4                & 8.9                  & 72.2              & 0.391                  & -                    \\ %
            CMA-ES   & 6.1                 & 5.8                  & 92.5              & \textbf{0.366}         & 38.5                 \\ %
            Zeroth   & 4.1                 & 4.3                  & \textbf{98.9}     & 0.397                  & \textbf{0.5}         \\ %
            DGS      & \textbf{3.9}        & \textbf{4.1}         & 97.2              & 0.385                  & 1.5                  \\ %
            \bottomrule
      \end{tabular}
      \caption{\textbf{ViperX, Single Trajectory}}
\end{table}

\begin{table}[H]
      \centering
      \begin{tabular}{lrrrrr}
            \toprule
            Approach & Pos Err             & Orn Err              & Survival          & Velocity               & Optimize Time        \\
                     & \it{cm}$\downarrow$ & \it{deg}$\downarrow$ & \it{\%}$\uparrow$ & \it{rad/s}$\downarrow$ & \it{sec}$\downarrow$ \\
            \midrule
            Random   & 17.0                & 9.4                  & 80.6              & 0.384                  & -                    \\ %
            CMA-ES   & 8.5                 & 7.0                  & 87.8              & 0.385                  & 615.7                \\ %
            Zeroth   & \textbf{3.4}        & \textbf{3.7}         & \textbf{97.8}     & \textbf{0.362}         & \textbf{7.3}         \\ %
            DGS      & 5.5                 & 4.2                  & \textbf{97.8}     & 0.394                  & 43.4                 \\ %
            \bottomrule
      \end{tabular}
      \caption{\textbf{ViperX, Multi Trajectory}}
\end{table}
\begin{table}[H]
      \centering
      \begin{tabular}{lrrrrr}
            \toprule
            Approach & Pos Err             & Orn Err              & Survival          & Velocity               & Optimize Time        \\
                     & \it{cm}$\downarrow$ & \it{deg}$\downarrow$ & \it{\%}$\uparrow$ & \it{rad/s}$\downarrow$ & \it{sec}$\downarrow$ \\
            \midrule
            Random   & 3.4                 & 8.4                  & 96.1              & 0.561                  & -                    \\ %
            CMA-ES   & \textbf{2.1}        & \textbf{5.9}         & \textbf{100.0}    & \textbf{0.376}         & 251.1                \\ %
            Zeroth   & 2.8                 & 8.0                  & 98.9              & 0.450                  & \textbf{1.1}         \\ %
            DGS      & 3.5                 & 9.0                  & 99.4              & 0.455                  & 9.1                  \\ %
            \bottomrule
      \end{tabular}
      \caption{\textbf{Quadruped, Single Trajectory}}
\end{table}
\begin{table}[H]
      \centering
      \begin{tabular}{lrrrrr}
            \toprule
            Approach & Pos Err             & Orn Err              & Survival          & Velocity               & Optimize Time        \\
                     & \it{cm}$\downarrow$ & \it{deg}$\downarrow$ & \it{\%}$\uparrow$ & \it{rad/s}$\downarrow$ & \it{sec}$\downarrow$ \\
            \midrule
            Random   & 2.8                 & 4.4                  & \textbf{99.6}     & 0.252                  & -                    \\ %
            CMA-ES   & \textbf{1.8}        & \textbf{2.4}         & \textbf{99.6}     & \textbf{0.188}         & 11505.2              \\ %
            Zeroth   & 1.9                 & 2.9                  & 99.1              & 0.271                  & 61.95                \\ %
            DGS      & 1.9                 & 2.9                  & 98.7              & 0.247                  & \textbf{53.5}        \\ %
            \bottomrule
      \end{tabular}
      \caption{\textbf{Bimanual, Multi Trajectory}}
\end{table}

\subsection{Tracking Size}
\label{sec:appendix:tracking_size}

\textbf{Reward Definition.}
To capture the notion of the robot's size $s_{achieved}$, we use the sum of the lengths of all of the robot's geometries, where the length of a geometry is the magnitude of its longest dimension.
Since different robot types have very different sizes, we use a target size $s_{target}$ and define the size penalty as:
\begin{equation}
      -T\alpha_{size} \min\left(
      s_{achieved}-s_{target},0
      \right)
      \label{eq:tracking_size}
\end{equation}
This term is zero only if the size of the robot is below the target size, and is added to \eqref{eq:tracking_only}.
We use $\alpha_{size}=0.1$ for all design spaces, while tuning $s_{target}$ to be 2.0m for the ViperX design space, 6.5m for the quadruped design space, and 5.0m for the mobile bimanual design space.

\textbf{Guidance Scale.}
Since diffusion training requires the model to look at its previous diffusion time step's noisy token to predict how to denoise, we find that the attention of a token to itself is significantly higher than its attention to other tokens.
This means the gradients of rewards that use embodiment tokens (e.g., size, weight) are also significantly higher than those of rewards that use dynamics tokens (e.g., tracking error).
We found that not accounting for this magnitude difference led to invalid robot outputs from diffusion.
To address this, we use $\alpha_{size}=0.005$ for the guidance term, but keep $\alpha_{size}=0.1$ for the actual reward reported.

\textbf{Reward-Specific Metrics.}
In addition to the tracking metrics (\S~\ref{sec:appendix:tracking_only}), we report the size of the robot as defined above.

\begin{table}[H]
      \centering
      \begin{tabular}{lrrrrr}
            \toprule
            Approach & Pos Err             & Orn Err              & Survival          & Size               & Optimize Time        \\
                     & \it{cm}$\downarrow$ & \it{deg}$\downarrow$ & \it{\%}$\uparrow$ & \it{m}$\downarrow$ & \it{sec}$\downarrow$ \\
            \midrule
            Random   & 19.4                & 8.9                  & 72.2              & 2.35               & -                    \\ %
            CMA-ES   & 6.2                 & 6.3                  & 94.4              & 2.21               & 38.3                 \\ %
            Zeroth   & \textbf{4.3}        & \textbf{4.3}         & \textbf{96.7}     & 2.31               & \textbf{0.5}         \\ %
            DGS      & 4.6                 & 4.4                  & 96.1              & \textbf{2.19}      & 1.9                  \\ %
            \bottomrule
      \end{tabular}
      \caption{\textbf{ViperX, Single Trajectory}}
\end{table}

\begin{table}[H]
      \centering
      \begin{tabular}{lrrrrr}
            \toprule
            Approach & Pos Err             & Orn Err              & Survival          & Size               & Optimize Time        \\
                     & \it{cm}$\downarrow$ & \it{deg}$\downarrow$ & \it{\%}$\uparrow$ & \it{m}$\downarrow$ & \it{sec}$\downarrow$ \\
            \midrule
            Random   & 17.0                & 9.4                  & 80.6              & 2.42               & -                    \\ %
            CMA-ES   & 7.4                 & 6.7                  & 91.1              & \textbf{2.28}      & 564.6                \\ %
            Zeroth   & 3.3                 & 3.8                  & 96.7              & 2.32               & \textbf{1.2}         \\ %
            DGS      & \textbf{2.8}        & \textbf{3.7}         & \textbf{98.9}     & 2.29               & 3.6                  \\ %
            \bottomrule
      \end{tabular}
      \caption{\textbf{ViperX, Multi Trajectory}}
\end{table}
\begin{table}[H]
      \centering
      \begin{tabular}{lrrrrr}
            \toprule
            Approach & Pos Err             & Orn Err              & Survival          & Size               & Optimize Time        \\
                     & \it{cm}$\downarrow$ & \it{deg}$\downarrow$ & \it{\%}$\uparrow$ & \it{m}$\downarrow$ & \it{sec}$\downarrow$ \\
            \midrule
            Random   & 3.4                 & 8.4                  & 96.1              & 9.25               & -                    \\ %
            CMA-ES   & \textbf{2.3}        & \textbf{7.2}         & \textbf{100.0}    & 7.28               & 263.3                \\ %
            Zeroth   & 4.2                 & 11.6                 & 94.4              & 7.40               & \textbf{3.2}         \\ %
            DGS      & 4.6                 & 12.6                 & 94.4              & \textbf{7.17}      & 9.0                  \\
            \bottomrule
      \end{tabular}
      \caption{\textbf{Quadruped, Single Trajectory}}
\end{table}
\begin{table}[H]
      \centering
      \begin{tabular}{lrrrrr}
            \toprule
            Approach & Pos Err             & Orn Err              & Survival          & Size               & Optimize Time        \\
                     & \it{cm}$\downarrow$ & \it{deg}$\downarrow$ & \it{\%}$\uparrow$ & \it{m}$\downarrow$ & \it{sec}$\downarrow$ \\
            \midrule
            Random   & 2.8                 & 4.4                  & 99.6              & 4.84               & -                    \\ %
            CMA-ES   & 1.8                 & \textbf{2.7}         & \textbf{100.0}    & 4.79               & 11505.2              \\ %
            Zeroth   & 1.7                 & \textbf{2.7}         & 99.1              & \textbf{4.71}      & \textbf{10.7}        \\ %
            DGS      & \textbf{1.6}        & 2.8                  & 99.6              & 4.73               & 58.9                 \\ %
            \bottomrule
      \end{tabular}
      \caption{\textbf{Bimanual, Multi Trajectory}}
\end{table}

\subsection{Tracking Weight}

\textbf{Reward Definition.}
The robot's total mass $m_{achieved}$ is the sum of all of its geometries' masses.
Like \eqref{eq:tracking_size}, we set a target mass $m_{target}$, and scale it by the number of timesteps:
\begin{equation}
      -T\alpha_{mass} \min\left(
      m_{achieved}-m_{target},0
      \right)
      \label{eq:tracking_weight}
\end{equation}
For the ViperX design space, $\alpha_{mass}$ is set to 10, with a target weight of 3.2kg.
This term is added to \eqref{eq:tracking_only}.

\textbf{Guidance Scale.}
Like in \S~\ref{sec:appendix:tracking_size}, we use $\alpha_{mass}=0.005$ to account for higher attention of tokens to themselves.

\textbf{Reward-Specific Metrics.}
In addition to the tracking-only metrics (\S~\ref{sec:appendix:tracking_only}), we report the total mass of the robot.

\begin{table}[H]
      \centering
      \begin{tabular}{lrrrrr}
            \toprule
            Approach & Pos Err             & Orn Err              & Survival          & Weight              & Optimize Time        \\
                     & \it{cm}$\downarrow$ & \it{deg}$\downarrow$ & \it{\%}$\uparrow$ & \it{kg}$\downarrow$ & \it{sec}$\downarrow$ \\
            \midrule
            Random   & 19.4                & 8.9                  & 72.2              & 3.2                 & -                    \\ %
            CMA-ES   & 13.2                & 6.3                  & 80.6              & 3.2                 & 41.8                 \\ %
            Zeroth   & 8.7                 & 6.3                  & 91.7              & 3.0                 & 1.0                  \\ %
            DGS      & 9.2                 & 6.1                  & 92.2              & 3.0                 & 2.9                  \\ %
            \bottomrule
      \end{tabular}
      \caption{\textbf{ViperX, Single Trajectory}}
\end{table}

\begin{table}[H]
      \centering
      \begin{tabular}{lrrrrr}
            \toprule
            Approach & Pos Err             & Orn Err              & Survival          & Weight              & Optimize Time        \\
                     & \it{cm}$\downarrow$ & \it{deg}$\downarrow$ & \it{\%}$\uparrow$ & \it{kg}$\downarrow$ & \it{sec}$\downarrow$ \\
            \midrule
            Random   & 17.0                & 9.4                  & 80.6              & 3.3                 & -                    \\ %
            CMA-ES   & 17.4                & 5.6                  & 73.3              & 3.2                 & 516.4                \\ %
            Zeroth   & 8.6                 & \textbf{5.1}         & \textbf{97.2}     & \textbf{3.1}        & \textbf{7.3}         \\ %
            DGS      & \textbf{8.1}        & 5.7                  & 91.7              & \textbf{3.1}        & 45.3                 \\ %
            \bottomrule
      \end{tabular}
      \caption{\textbf{ViperX, Multi Trajectory}}
\end{table}

\section{Additional Method Details}

\begin{table}[t]
      \footnotesize
      \centering
      \begin{tabular}{llrl}
            \toprule
            RoboToken Type      & Attribute          & Dim & Encoding   \\
            \midrule
            Link                & Geometry Type      & 3   & Binary     \\
                                & Geometry Size      & 3   & None       \\
                                & Mass               & 1   & Log        \\
                                & Inertia Position   & 3   & None       \\
                                & Inertia Rotation   & 9   & None       \\
                                & Diagonal Inertia   & 3   & Log        \\
                                & Friction           & 3   & None       \\
                                & Contact Dim.       & 3   & Binary     \\
                                & Color              & 4   & None       \\
                                & Free Link ID       & 1   & Binary     \\
                                & Track Link ID      & 1   & Binary     \\
                                & Link ID            & -   & Pos Emb    \\
            \midrule
            Dynamic Joint       & Joint Type         & 2   & Binary     \\
                                & Range              & 2   & None       \\
                                & Armature           & 1   & Log        \\
                                & Damping            & 1   & Log        \\
                                & Friction Loss      & 1   & Log        \\
                                & Stiffness          & 1   & Log        \\
                                & Spring Reference   & 1   & None       \\
                                & Position Reference & 1   & None       \\
                                & Position to Link   & 3   & None       \\
                                & Rotation  to Link  & 9   & None       \\
                                & Link ID            & 7   & Binary     \\
                                & Dyna Joint ID      & -   & Pos Emb    \\
            \midrule
            Fixed Joint         & Position           & 3   & None       \\
                                & Rotation           & 9   & None       \\
                                & Link ID            & 7   & Binary     \\
                                & Fixed Joint ID     & -   & Pos Emb    \\
            \midrule
            Actuator            & Control Range      & 2   & None       \\
                                & Force Range        & 4   & Signed Log \\
                                & Position Gain      & 1   & Log        \\
                                & Velocity Gain      & 1   & Log        \\
                                & Gear Ratio         & 1   & None       \\
                                & Actuator Type      & 1   & Binary     \\
                                & Dyna Joint ID      & 6   & Binary     \\
                                & Actuator ID        & -   & Pos Emb    \\
            \midrule
            Free/Track Link Obs & Position           & 3   & None       \\
                                & Rotation           & 9   & None       \\
                                & Free/Track Link ID & -   & Pos Emb    \\
                                & Time ID            & -   & Pos Emb    \\
            \midrule
            Dynamic Joint Obs   & Joint Pos          & 1   & None       \\
                                & Joint Vel          & 1   & None       \\
                                & Dyna Joint ID      & -   & Pos Emb    \\
                                & Time ID            & -   & Pos Emb    \\
            \midrule
            Actuator Obs        & Velocity           & 1   & None       \\
                                & Force              & 1   & None       \\
                                & Actuator ID        & -   & Pos Emb    \\
                                & Time ID            & -   & Pos Emb    \\
            \midrule
            Control             & Target Pos         & 1   & None       \\
                                & Actuator ID        & -   & Pos Emb    \\
                                & Time ID            & -   & Pos Emb    \\
            \midrule
            Target Pose         & Position           & 3   & None       \\
                                & Rotation           & 9   & None       \\
                                & Track Link ID      & -   & Pos Emb    \\
                                & Time ID            & -   & Pos Emb    \\
            \bottomrule
      \end{tabular}
      \caption{\footnotesize\textbf{RoboToken Schema.} Not listed here are ball joints, whose schema is identical to that of dynamic joints', but whose observations are quaternions and angular velocities instead.}
      \label{tab:robotoken_schema}
\end{table}

\subsection{RoboToken}

\textbf{Token Count Comparison.}
The text token count in Figure 4 in the main paper is calculated from the robot's MJCF format tokenized with \href{https://platform.openai.com/tokenizer}{GPT-4o's tokenizer}.
We compared against text tokens as these are the only other complete robot representation that can be used directly both as input to and output from a neural network.

\textbf{Inertia Splitting.}
To maintain consistency in \textit{RoboToken}, we require that all inertial properties arise strictly from the link's geometry primitives.
When a user-provided robot description contains a link with multiple geometries but a single lumped inertia tensor, we apply a heuristic algorithm to split this lumped inertia into individual geometry inertias.
\begin{enumerate}
      \item We assume uniform density $\rho$ across all primitives within a link.
      \item For each primitive $i$, we calculate its local inertia $I_i$ and mass $m_i$ based on its geometry (e.g., box, sphere, or cylinder).
      \item We apply the Parallel Axis Theorem~\cite{wiki:Parallel_axis_theorem} to shift these inertias to the link's common center of mass:
            $$I_{\text{total}} = \sum (I_i + m_i [(\mathbf{r}_i \cdot \mathbf{r}_i) \mathbf{E} - \mathbf{r}_i \otimes \mathbf{r}_i])$$
            where $\mathbf{r}_i$ is the displacement vector from the primitive center to the link center.
\end{enumerate}

\textbf{Transform Canonicalization.}
To ensure that the model does not have to learn redundant spatial offsets, we preprocess all MJCF/URDF files to collapse all transforms into the joint token's transforms.
This means all geometries are effectively at the origin of their link's frame.

\textbf{Attribute Encoding.}
Robot embodiment attributes can span different data types (boolean, enums, integers, continuous values) and ranges.
For all boolean/enum/integer attributes (such as joint types, geometry types, or ID attributes), we use binary encoding.
For continuous-valued attributes that range over many orders of magnitude (such as inertia, motor gains), we use log or signed log encoding.
We include the RoboToken schema along with each attribute's encoding in Table~\ref{tab:robotoken_schema}.

\textbf{Context Length Considerations.}
Since self-attention is $O(n^2)$ in sequence length, carefully determining the model's context length is critical for training and inference speeds.
Therefore, we remove all link pose observations except for two types of links.
First are free links in free-base robots (e.g., quadrupeds).
Second are end-effectors, which are referred to in the schema as ``Track Link'' (Table~\ref{tab:robotoken_schema}) to allow generalization to other robot body parts.
Since a robot contains many more links than are usually task-relevant, omitting all these other links significantly reduces the context length for the dynamics token portion of input tokens.
For each design space, we train using that design space's max sequence length.

\begin{table}[t]
      \centering
      \begin{tabular}{rl}
            \toprule
            Hyperparameter      & Value
            \\
            \midrule
            Hidden dim          & 256              \\
            Num layers          & 8                \\
            Num heads           & 4                \\
            Dropout             & 0.0              \\
            LR                  & $5\times10^{-3}$ \\
            LR ramp up          & 500 steps        \\
            Batch size          & 64               \\
            Weight decay        & 0.0              \\
            EMA power           & 0.75             \\
            Num epochs          & 50               \\
            Num steps per epoch & 16,384           \\
            \bottomrule
      \end{tabular}
      \caption{\textbf{Model Hyperparameters}}
      \label{tab:model_hparams}
\end{table}

\subsection{Transformer Transformer}
\label{sec:model_details}

\textbf{Hyperparameters.}
Our model and training hyperparameters are listed in Tab.~\ref{tab:model_hparams}.
We found that when training a unified model, a larger model trained significantly faster. We opted for doubling the hidden dimension and number of heads, and increasing the number of layers to 12 (large model, Fig.~\ref{fig:bimanual_control}).
To help this larger model train stably, we use a smaller learning rate of $1\times10^{-4}$.

\textbf{Joint Token Ordering.}
We represent each joint with two embodiment tokens, with each part responsible for pointing to one link and encoding the transform to that link.
All other attributes of the joint are duplicated between the two tokens.
To help the diffusion model distinguish each joint's two tokens, we add a joint ordering ID attribute, which is 1 for the token with the higher link ID.
Joint order IDs are converted into a learnable positional embedding and added to the joint tokens.

\textbf{Dynamics Model Temporal Resolution.}
We experimented with including more episode timesteps in the model context, hypothesizing that increased temporal resolution would allow the model to learn a more accurate dynamics model.
However, these models did not significantly improve downstream hardware optimization performance, yet they significantly increased training and inference time.
Thus, we opted for 8 time steps, which we empirically observed to strike the best balance between speed and performance for our experiments.

\textbf{Diffusion Process Length.}
We also found that while modeling dynamics benefits from longer diffusion processes, modeling actions with shorter diffusion processes was sufficient.
Specifically, precise state predictions for downstream reward-guided diffusion, especially for the tracking error objective, are critical for correct guidance, and a longer diffusion process achieves lower prediction error.
Meanwhile, action prediction is robust to small deviations from the expert.
Thus, we use 100 and 5 time steps for motion-to-robot and cross-embodiment control, respectively.

\textbf{Explicit Padding Tokens.}
For motion-to-robot, the model can generate sequences with variable lengths depending on the number of links, joints, and actuators of the robot it will generate, so the model is trained (without masking in the attention mechanism) to explicitly predict padding tokens, which are special-valued tokens detectable by our detokenizer.
Meanwhile, for cross-embodiment control, all padding tokens are masked out.

\textbf{Checkpoint Selection.}
We use the best checkpoints, which are typically among the last few checkpoints.
We found that although motion-to-robot optimization performance improves quickly early in training, cross-embodiment control performance plateaus only in later checkpoints.

\subsection{Data}
\label{sec:data_details}

\subsubsection{\textbf{Procedural Generation}}\label{sec:procedural_generation}
We define a procedural generation grammar that grows robots from a common root component outwards.
We extend the MuJoCo MJCF XML format to support a meta-level language in the custom data XML fields and inject a parser for handling such meta-level fields.
This additional meta-level functionality allows users to define randomized continuous attributes for a subcomponent (e.g., dimension of a leg), constraints between attributes (e.g., length of left calf should equal length of right calf), as well as randomized discrete attributes (e.g., rotation axis of a joint, inclusion of a spring-loaded leg linkage mechanism).
Thus, given a procedural robot design space, each robot embodiment can be parameterized by a list of discrete choices and a list of continuous choices, where the length of both lists is variable (some discrete choices can alter the degrees of freedom of the robot).
Using our system, existing robot definitions can be easily extended to support a wide range of procedural geometry, kinematics, and dynamics variation, generating the source of embodiment diversity in our dataset.

\textbf{ViperX.} For discrete variations, we include 4 different mounting orientations (left, right, upright, downward), 3 different DoF variations (5 DoF, 6 DoF, and 7 DoF), and 3 joint rotation axis variations per upper joint (X-axis, Y-axis, and Z-axis rotation). For continuous variations, we include mounting positions (within 0.7m, 1.0m, and 0.6m for XYZ directions respectively), mounting Z-angle ($-\pi/2\,\text{rad}$ to $+\pi/2\,\text{rad}$), and link length variations ($-5\,\text{cm}$ to $+20\,\text{cm}$ relative increase).

\textbf{Quadruped Manipulator.}
For discrete variations, we include 2 arm mount options (fixed or sliding linear rail along the back), 2 leg design options per leg (spring-loaded linkage or serial leg, chosen independently for all 4 legs), and knee direction (flipped forward vs. flipped backward, shared within the front pair and within the back pair), yielding 7 binary design choices in total.
For continuous variations, we include arm link length variations ($-10\,\text{cm}$ to $40\,\text{cm}$ relative to initial length), arm mounting position ($-20\,\text{cm}$ to $+20\,\text{cm}$), body length ($10\,\text{cm}$ to $50\,\text{cm}$), leg link lengths ($20\,\text{cm}$ to $60\,\text{cm}$), leg mounting Z-angle per leg ($-0.7\,\text{rad}$ to $0.7\,\text{rad}$), battery placement along the length of the robot (heavy green box at the core of the robot, $-20\,\text{cm}$ to $+20\,\text{cm}$).

\textbf{Mobile Bimanual.}
For discrete variations, we include three spine design choices (a fixed spine, a telescopically sliding spine, and a bending spine) and a torso $45^\circ$ pitching DoF (included or not, allowing the robot to lean forward).
For continuous variations, we include spine length variations ($60\,\text{cm}$ to $140\,\text{cm}$ for the fixed spine, $10\,\text{cm}$ to $50\,\text{cm}$ for the telescopically sliding spine, and $20\,\text{cm}$ to $80\,\text{cm}$ for the bending spine), shoulder offset ($4\,\text{cm}$ to $10\,\text{cm}$ in the y direction, $-3\,\text{cm}$ to $8\,\text{cm}$ in the z direction), and arm link lengths ($2\,\text{cm}$ to $8\,\text{cm}$ for the shoulder link, $1\,\text{cm}$ to $18\,\text{cm}$ for the arm and forearm).

\textbf{ALOHA.}
For discrete variations, we include three mounting point orientations (upright facing each other, vertically mounted on either side, or upside down).
For continuous variations, we include mounting position variation ($-40\,\text{cm}$ to $0\,\text{cm}$ for x, $40\,\text{cm}$ to $60\,\text{cm}$ for inter-arm distance, and $60\,\text{cm}$ to $90\,\text{cm}$ for the z height), as well as arm length variations ($10\,\text{cm}$ to $50\,\text{cm}$ for the arm extension, $5\,\text{cm}$ to $25\,\text{cm}$ for the forearm extension).
We constrain the bimanual set up to be symmetric about the XZ plane.

\subsubsection{\textbf{Whole-Body Controller (WBC) Training}}
Following the WBC controller formulation from UMI on Legs~\cite{ha2024umilegs}, we train our WBC using PPO~\cite{schulman2017proximal}.
This follows a standard RL controller training procedure that uses a task reward (tracking in our case,~\eqref{eq:tracking_only}) with regularization penalty terms (joint velocity, acceleration, limits, body orientation, etc.) and in-training perturbations (random kicks, transports) to make the policy robust.
In addition to the robot's states, these RL experts also observe the continuous variation parameters of the robot's embodiment (\S~\ref{sec:procedural_generation}) and 4 target poses into the future.
At the beginning of each episode, a random trajectory from the training set of end-effector trajectories is selected, a random transform augmentation is applied, and the robot is then controlled to track the target trajectories.
Episodes are terminated if the robot falls over, with a termination penalty applied.
We train one RL expert per discrete choice, which for the 7 binary design choices in the quadruped case yields 128 specialized experts.
Training each RL policy takes 16 hours on an NVIDIA A100.
At data generation time, we procedurally generate robots from a stream of randomized discrete and continuous choices, load the corresponding RL policy, and control the robot to track a training trajectory.
Each rollout is then tokenized into a RoboTokens dataset.
We add control perturbations to the expert actions to increase state diversity, but log the unperturbed expert action for supervision~\cite{truong2024pdp}.

\subsubsection{\textbf{Data Quality}}
We ensure high tracking performance in training datasets in two ways.
First, we filter all episodes that last less than 100 time steps (i.e., 2 seconds), which represents embodiments that fell over (for quadrupeds) or exceeded the tracking position error termination threshold during that time.
Second, we bias the data generation towards higher-performance embodiments for the ViperX and ALOHA design space.
While tracking error is reasonable for the quadruped and mobile bimanual design spaces, the tracking error of the random ViperX and ALOHA embodiment is very high ($19.4\,\text{cm}$, $8.9^\circ$, Table~\ref{tab:viperx:tracking_only}) since the kinematics of fixed-base arms critically determine whether a target pose is even reachable.
To improve training embodiment quality, we run CMA-ES on the training trajectories with a population of 5 for 3 generations and use all embodiments sampled during CMA-ES as training data.
Although improving training data quality is one way CMA-ES and our approach can be paired together, we believe their integration could also occur at inference time, where our algorithm acts as the sampler for an evolutionary algorithm.
We leave this investigation for future work.

\subsubsection{\textbf{Dataset Size}}
Our ViperX dataset includes 3.8M episodes with 2B time steps in total.
Our quadruped dataset includes 1.3M episodes with 500M time steps in total.
Finally, our bimanual mobile dataset includes 50K episodes and 69M time steps in total.
In each dataset, each robot embodiment design is used for 10 episodes, so the total number of training embodiments is 380K, 130K, and 5K for the ViperX, quadruped, and bimanual design spaces respectively.

\subsection{Inference}

\textbf{Stochastic DDIM sampling.}
DDIM~\cite{songdenoising} has a hyperparameter for controlling the amount of noise injected into the sampling procedure, typically denoted as $\eta$ in DDIM implementations.
Although deterministic DDIM sampling ($\eta=0.0$) is common, we found that using stochastic sampling with $\eta=1.0$ for Dynamics Self-Guidance (DGS) specifically improved motion-to-robot optimization performance significantly.
Meanwhile, tuning $\eta$ did not affect our Zeroth Order sampler's performance.
This suggests that, like all first-order methods, our DGS sampler is prone to local minima, but injecting a gradually decaying noise term aids exploration along the diffusion process.

\textbf{Guidance Scale.}
For DGS, we use the highest guidance scale that does not push the diffusion model out of distribution to generate invalid robots.
It is set to 50, 100, and 0.2 for the ViperX, quadruped, and bimanual design spaces respectively,
and raised to 500 for the multi-trajectory ViperX experiments.

\end{document}